%% file: iclr2027_conference.tex
\documentclass{article} 
\usepackage{iclr2027_conference,times}

\input{math_commands.tex}

\usepackage{hyperref}
\usepackage{url}
\usepackage[utf8]{inputenc} 
\usepackage[T1]{fontenc}    
\usepackage{booktabs}       
\usepackage{amsfonts}       
\usepackage{nicefrac}       
\usepackage{microtype}      
\usepackage{xcolor}         

\usepackage{pifont}
\usepackage{colortbl}
\usepackage{graphicx}
\usepackage{multirow}
\usepackage{subcaption}
\usepackage{threeparttable}
\usepackage[percent]{overpic}  
\usepackage{makecell}
\usepackage{booktabs}

\usepackage{listings}
\usepackage{tcolorbox}
\usepackage{fancyvrb}
\usepackage{fvextra}
\usepackage{wrapfig}
\usepackage{algorithm}
\usepackage{algpseudocode}

\newcommand{\secondbest}[1]{\underline{#1}}
\newcommand{\best}[1]{\textbf{#1}}
\DeclareSymbolFont{extraup}{U}{zavm}{m}{n}
\DeclareMathSymbol{\newcrossmark}{\mathalpha}{extraup}{129}
\definecolor{softgreen}{RGB}{46,160,67}
\definecolor{lightgreen}{RGB}{234,241,223}
\definecolor{lightred}{RGB}{252,234,220}
\definecolor{lightblue}{RGB}{216,230,231}
\definecolor{lightbrown}{RGB}{239,227,211}
\definecolor{lightpurple}{RGB}{198,213,234}
\definecolor{lightdark}{RGB}{223,218,212}
\newcommand{\cmark}{\textcolor{softgreen}{\ding{51}}}
\newcommand{\xmark}{\textcolor{red}{\ding{55}}}
\begingroup

\footnotetext{Equal contribution.}
\endgroup

\title{ReVA: A Scene-Centric Dataset Beyond Repetition for Remote Sensing Video Question Answering}

\author{
  \textbf{Zhen Yao$^{1*}$, Likai Wang$^{1*}$, Yuming Yang$^{1}$, Zhihao Zheng$^{1}$, Bo Lang$^{1}$, Qiuyu Tang$^{1}$,} \\
  \textbf{Jialu Sheng$^{1}$, Jingqi Xu$^{2}$, Yuehai Yang$^{1}$, Jumal Barker$^{1}$, Xiaowen Ying$^{3}$, Mooi Choo Chuah$^{1}$}\\
  $^{1}$Lehigh University, $^{2}$University of Southern California, $^{3}$Qualcomm AI Research\\
}

\iclrfinalcopy 
\begin{document}

\maketitle

\input{sec/0_abstract}    
\input{sec/1_intro}
\input{sec/2_related}
\input{sec/3_method}
\input{sec/4_result}


\bibliography{iclr2027_conference}
\bibliographystyle{iclr2027_conference}

\input{sec/X_suppl}

\end{document}

%% file: math_commands.tex
\usepackage{amsmath,amsfonts,bm}

\def\eqref#1{equation~\ref{#1}}

\def\1{\bm{1}}

\DeclareMathAlphabet{\mathsfit}{\encodingdefault}{\sfdefault}{m}{sl}
\SetMathAlphabet{\mathsfit}{bold}{\encodingdefault}{\sfdefault}{bx}{n}



%% file: sec/0_abstract.tex
\begin{abstract}
Multimodal Large Language Models (MLLMs) have demonstrated remarkable advances in remote sensing. However, existing remote sensing multimodal reasoning benchmarks exhibit two critical limitations: they rely on (i) template-driven questions, which causes repetitive questions; and (ii) static images that fail to capture the inherent temporal nature of drone/UAV videos. 
This leaves systematic evaluation of remote sensing video reasoning largely unexplored. To address this gap, we introduce \textbf{ReVA}, a new dataset for remote sensing video question answering, designed to assess spatiotemporal, scene-centric, and reasoning-oriented capabilities of MLLMs. ReVA comprises 2,438 drone videos spanning 18 cities worldwide (580K frames) and 22K high-quality question–answer pairs across 11 challenging QA tasks. We develop a semi-automatic annotation pipeline that leverages Text LLMs and MLLMs for question-answer generation with human verification. We evaluate 23 proprietary and open-source Video LLMs on ReVA, exposing fundamental limitations of current models. These findings position ReVA as a critical benchmark toward better remote sensing video understanding and temporal reasoning capabilities for real-world deployments. Our code and dataset are available at: https://github.com/zyaocoder/ReVA
\end{abstract}

%% file: sec/1_intro.tex
\vspace{-1.2\baselineskip}
\section{Introduction}  \label{sec:intro}
\vspace{-0.3\baselineskip}
Recent advances in Multimodal Large Language Models (MLLMs) have significantly improved remote sensing image understanding, achieving strong performance in scene parsing. By coupling visual representation with natural language, these approaches demonstrate enhanced reasoning capabilities for complex aerial scenes. \par

Despite these advances, current remote sensing scene understanding faces two critical limitations: (i) \textbf{Template-driven Questions}: Existing benchmarks \cite{wang2024earthvqa,massih2026reasoning} rely on pre-defined, fixed question templates, e.g., "Is there \texttt{[Class]} in the image?". While improving annotation scalability, it restricts linguistic diversity and reasoning complexity, yielding an oversimplified, repetitive QA paradigm ill-suited for real-world deployment. (ii) \textbf{Limited Video-Centric Evaluation}: Most established works \cite{bashmal2023visual, zhang2023multistep} primarily focused on static images, yet remote sensing increasingly utilizes Unmanned Aerial Vehicles (UAVs) to capture continuous videos where the scenes change rapidly. The inherent temporal dynamics pose fundamental challenges that image-based reasoning cannot fully address. Concurrent studies have begun extending aerial understanding to videos; however, their objectives are mostly navigation-related reasoning, leaving scene-centric VideoQA largely underexplored. \par

\begin{figure*}[!t]
\centering
\includegraphics[width=1\linewidth]{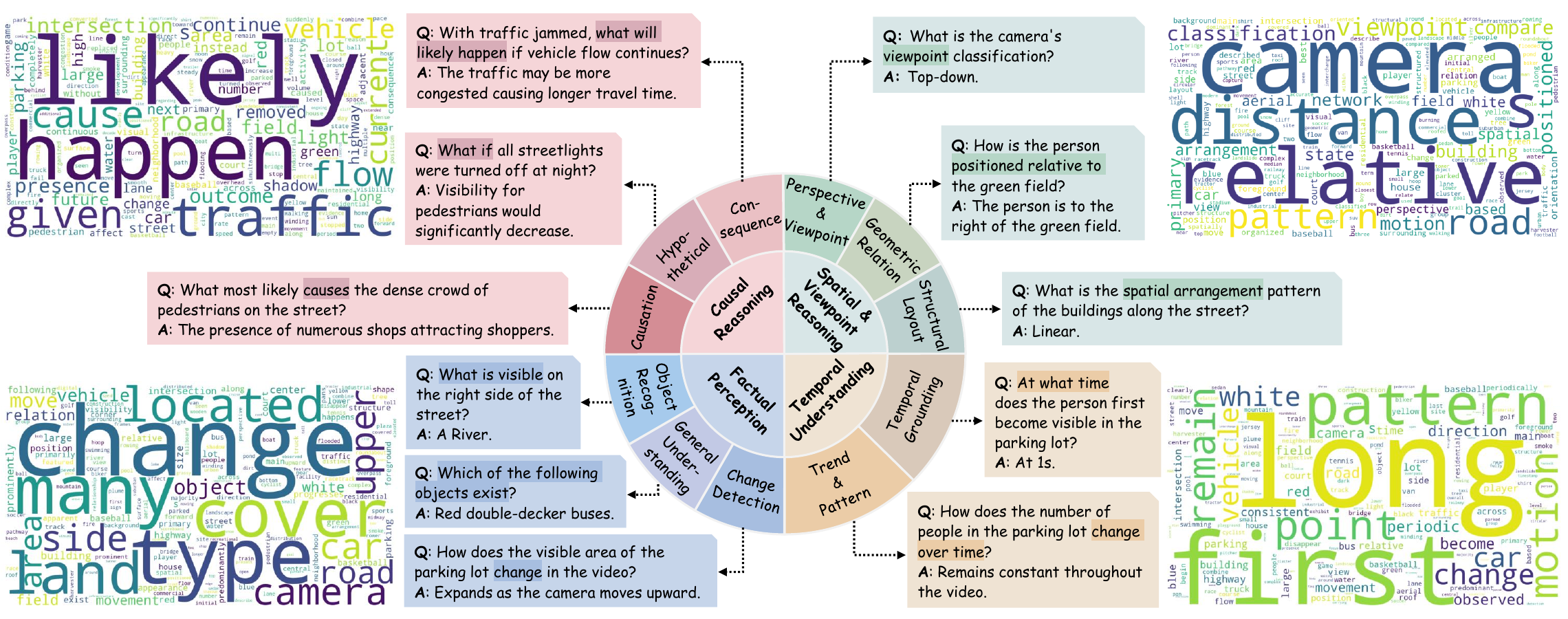}
\vspace{-1.4\baselineskip}
\caption{\textbf{Overview of ReVA.} ReVA contains 22K question-answer pairs under 4 major categories and 11 QA tasks. Word clouds are presented, highlighting different attributes of various categories. }
\label{fig:motivation}
\vspace{-1.2\baselineskip}
\end{figure*}

To this end, we introduce \textbf{ReVA}, a scene-centric dataset for remote sensing video understanding. Unlike prior image-based datasets, ReVA is built upon UAV video sequences and emphasizes holistic scene reasoning. As shown in Fig. \ref{fig:motivation}, it comprises diverse aerial scenarios and a wide range of question types, establishing a rigorous benchmark for thoroughly evaluating multimodal reasoning in real-world remote sensing contexts. Beyond standard question categories (e.g., factual perception), it introduces two unique yet challenging categories: temporal understanding and causal reasoning. Furthermore, we propose a five-stage semi-automatic QA generation and annotation workflow, designed to be generalizable and to facilitate future research in remote sensing video understanding. \par
 
Building on ReVA, we further propose \textbf{ReMoSense}, a motion-aware framework for remote sensing video understanding. ReMoSense explicitly captures camera motions and object temporal dynamics to achieve robust cross-frame alignment and long-range temporal reasoning. Specifically, we introduce global motion tokens that encode camera ego-motion via cost volumes, and object motion tokens that auto-regressively refine object temporal dynamics across frames. \par

In summary, our contributions in this paper include:
\begin{itemize}
\item We introduce ReVA, a fully real-world, scene-centric remote sensing video question answering dataset with \textbf{2,438} UAV videos (\textbf{580K} frames) from \textbf{18} cities worldwide and \textbf{22K} high-quality question-answer pairs (\textbf{17K} unique questions). Unlike prior UAV-view VideoQA benchmarks focused on urban/navigation settings, ReVA covers urban and rural scenes and evaluates a broader spectrum of remote sensing capabilities.

\item We propose ReMoSense, a motion-aware framework for remote sensing video understanding that explicitly disentangles camera motion and object temporal dynamics via two complementary components: global motion tokens encoding camera ego-motion through cost volumes, and object motion tokens that iteratively refine object dynamics across frames.

\item We develop a five-stage semi-automatic QA generation workflow for diverse, scene-centric, and reasoning-intensive QA pairs of \textbf{11} distinct tasks, reducing template-induced repetition.

\item We comprehensively evaluate 23 mainstream MLLMs on remote sensing video understanding capabilities. These in-depth analysis demonstrate limitations of existing models and heuristic observations for future works.
\end{itemize}

%% file: sec/2_related.tex
\vspace{-1\baselineskip}
\section{Related Work}  \label{sec:related}
\vspace{-0.6\baselineskip}
\subsection{Dataset and Benchmark}\label{subsec:dataset}
\vspace{-0.3\baselineskip}
In remote sensing (RS), VisualQA has attracted increasing research attention, yet existing \textbf{remote sensing VisualQA} datasets are image-based. HRVQA \cite{li2024hrvqa} releases a large-scale remote sensing VisualQA benchmark for specific challenges like scale variation. EarthVQA \cite{wang2024earthvqa} targets relation-centric reasoning by exploring geographical interactions among entities. RSVLM-QA \cite{zi2025rsvlm} emphasizes quantitative reasoning (e.g., counting). Despite these advances, image-based VisualQA cannot capture temporal dynamics, limiting evaluation of temporal reasoning required by applications such as disaster monitoring. Recent works have begun exploring dynamic UAV videos. UAVBench \cite{ferrag2026uavbench} studies UAV cognition through single- and two-frame based flight scenarios but still has no video inputs. Concurrently, RSVideo-10K \cite{zhou2026rsvideo} explores remote sensing video understanding. As RSVideo-10K is concurrent with our work, we discuss it for completeness rather than include it in the direct comparison in Table \ref{tab:datasetcomp}. These efforts demonstrate growing interest in video reasoning, while differing from ReVA in task formulation, scope, and repetition. ReVA focuses specifically on scene-centric VideoQA over real-world videos, with diverse questions spanning four major categories. \par

In contrast, \textbf{natural domain VideoQA} increasingly push beyond image-level reasoning toward long-horizon video understanding. NExT-QA \cite{xiao2021next} advances video understanding from shallow descriptions to deeper explanation of temporal action reasoning. LongVideoBench \cite{wu2024longvideobench} first extends the task to challenging long-context videos and constructs a dataset with hour-level videos, while EgoSchema \cite{mangalam2023egoschema} evaluates long-form comprehension on three-minute-long video clips. Ego4D \cite{graumanEgo4DWorld30002022} provides a large-scale egocentric video dataset containing daily human activities across diverse environments including home and workplace. MovieChat-1K \cite{song2024moviechat} evaluates long-video understanding with redundant frames to measure robustness to long-range dependencies. \par

As summarized in Table \ref{tab:datasetcomp}, existing remote sensing VisualQA and natural-domain VideoQA benchmarks leave a critical gap: a benchmark for remote sensing video understanding that demands RS-relevant spatiotemporal reasoning. This motivates the construction of ReVA, which addresses this gap along two key dimensions. First, ReVA emphasizes linguistic diversity, containing 17K unique questions that substantially reduce template-driven repetition. Second, ReVA introduces challenging question types (e.g., temporal understanding) absent from prior datasets. It enables systematic evaluation of video-specific reasoning under real-world remote sensing conditions. \par

\begin{table}[t]
\centering
\caption{\textbf{Dataset Comparison.} ReVA is a real-world, scene-centric remote sensing VideoQA dataset comprising 2.4K videos with 22K human-annotated QA pairs and 17K unique questions.}
\vspace{-0.6\baselineskip}
\resizebox{\textwidth}{!}{
\begin{tabular}{l|cccccccccccc}
\toprule
\multirow{2}{*}{Dataset} & \multirow{2}{*}{Year} & \multirow{2}{*}{\#Video} & \multirow{2}{*}{\#Image} & \multirow{2}{*}{\#QA} & \multirow{2}{*}{Annotate} & Unique & \multirow{2}{*}{FP} & \multirow{2}{*}{TU} & \multirow{2}{*}{SR} & \multirow{2}{*}{CR} \\
 & & & & & & question & & & & \\ 
\midrule
\rowcolor{lightblue} \multicolumn{11}{l}{\textbf{Natural Domain VideoQA}} \\
ActivityNet-QA \cite{yu2019activitynet} & 2019 & 5,800 & - & 58,000 & Manual & 18,897 & \cmark & \xmark & \cmark & \xmark\\
Social-IQ \cite{zadeh2019social} & 2019 & 1,250 & - & 7,500 & Manual & 5,713 & \cmark & \xmark & \xmark & \cmark \\
NExT-QA \cite{xiao2021next} & 2021 & 5,440 & - & 52,044 & Auto & 31,173 & \cmark & \cmark & \xmark & \cmark \\
WildQA \cite{castro2022wild} & 2022 & 369 & - & 916 & Manual & 251 & \cmark & \cmark & \cmark & \xmark \\
EgoSchema \cite{mangalam2023egoschema} & 2023 & 5,063 & - & 5,063 & Auto & 5,031 & \cmark & \cmark & \xmark & \cmark \\
MVBench \cite{li2024mvbench} & 2023 & 3,641 & - & 4,000 & Auto & 4,000 & \cmark & \cmark & \cmark & \cmark \\
STAR \cite{wu2024star} & 2024 & 23,517 & - & 60,000 & Auto & 2,385 & \cmark & \cmark & \xmark & \xmark \\
\midrule
\rowcolor{lightgreen} \multicolumn{11}{l}{\textbf{Remote Sensing VisualQA}} \\
RSIVQA \cite{zheng2021mutual} & 2021 & - & 37,264 & 111,134 & Auto & 91 & \cmark & \xmark & \xmark & \xmark \\
FloodNet \cite{rahnemoonfar2021floodnet} & 2021 & - & 3,200 & 11,000 & Manual & 15 & \cmark & \xmark & \xmark & \xmark \\
TextRS-VQA \cite{bashmal2023visual} & 2023 & - & 2,144 & 6,245 & Manual & 3,608 & \cmark & \xmark & \xmark & \xmark \\
CRSVQA \cite{zhang2023multistep} & 2023 & - & 4,639 & 4,644 & Manual & 674 & \cmark & \xmark & \cmark & \xmark \\
RSIEval \cite{hu2025rsgpt} & 2023 & - & 100 & 943 & Manual & 382 & \cmark & \xmark & \cmark & \xmark \\
EarthVQA \cite{wang2024earthvqa} & 2024 & - & 6,000 & 208,593 & Auto & 28 & \cmark & \xmark & \cmark & \xmark \\ 
SQuID \cite{massih2026reasoning} & 2026 & - & 2,000 & 2,000 & Manual & 391 & \cmark & \xmark & \cmark & \xmark \\
UAVBench \cite{ferrag2026uavbench} & 2026 & - & 47,488 & 47,488 & Auto & 47,453 & \cmark & \xmark & \cmark & \cmark \\
\midrule
ReVA (Ours) & 2026 & 2,438 & - & 21,773 & Manual & 16,695 & \cmark & \cmark & \cmark & \cmark\\
\bottomrule
\end{tabular}
}
\begin{tablenotes}[flushleft]
\scriptsize
\item * FP: Factual Perception, TU: Temporal Understanding, SR: Spatial Reasoning, CR: Causal Reasoning.
\end{tablenotes}
\label{tab:datasetcomp}
\vspace{-1.6\baselineskip}
\end{table}


\vspace{-0.6\baselineskip}
\subsection{Multimodal Reasoning in Remote Sensing} \label{subsec:remote sensing}
\vspace{-0.6\baselineskip}
Multimodal reasoning in remote sensing aims to integrate visual and language reasoning over aerial or satellite imagery. Some researchers design domain-specific MLLM. GeoChat \cite{kuckreja2024geochat} finetunes LLaVA \cite{liu2023visual,li2024llava,zhang2024llava} for a remote sensing MLLM on self-collected dataset. RSGPT \cite{hu2025rsgpt} finetunes the Q-Former network of LLMs for cross-modal alignment. Another line of works explore prompt- or reasoning-driven approaches. Prompt-RSVQA \cite{chappuis2022prompt} converts the image context into a text prompt for better remote sensing understanding. MQVQA \cite{zhang2023multistep} uses a question-driven multi-step reasoning mechanism to select local regions for fine-grained, question-related visual features. RemoteReasoner \cite{yao2025remotereasoner} aggregates pixel-, object-, and region-level for multi-scale reasoning. SkyAnchor \cite{sun2026memory} propsoes a Semantics-Aware Token Router to preserve tiny objects in video streams. \par

%% file: sec/3_method.tex
\vspace{-0.6\baselineskip}
\section{ReVA Dataset} \label{sec:dataset}

\begin{figure*}[!t]
\centering
\includegraphics[width=1\linewidth]{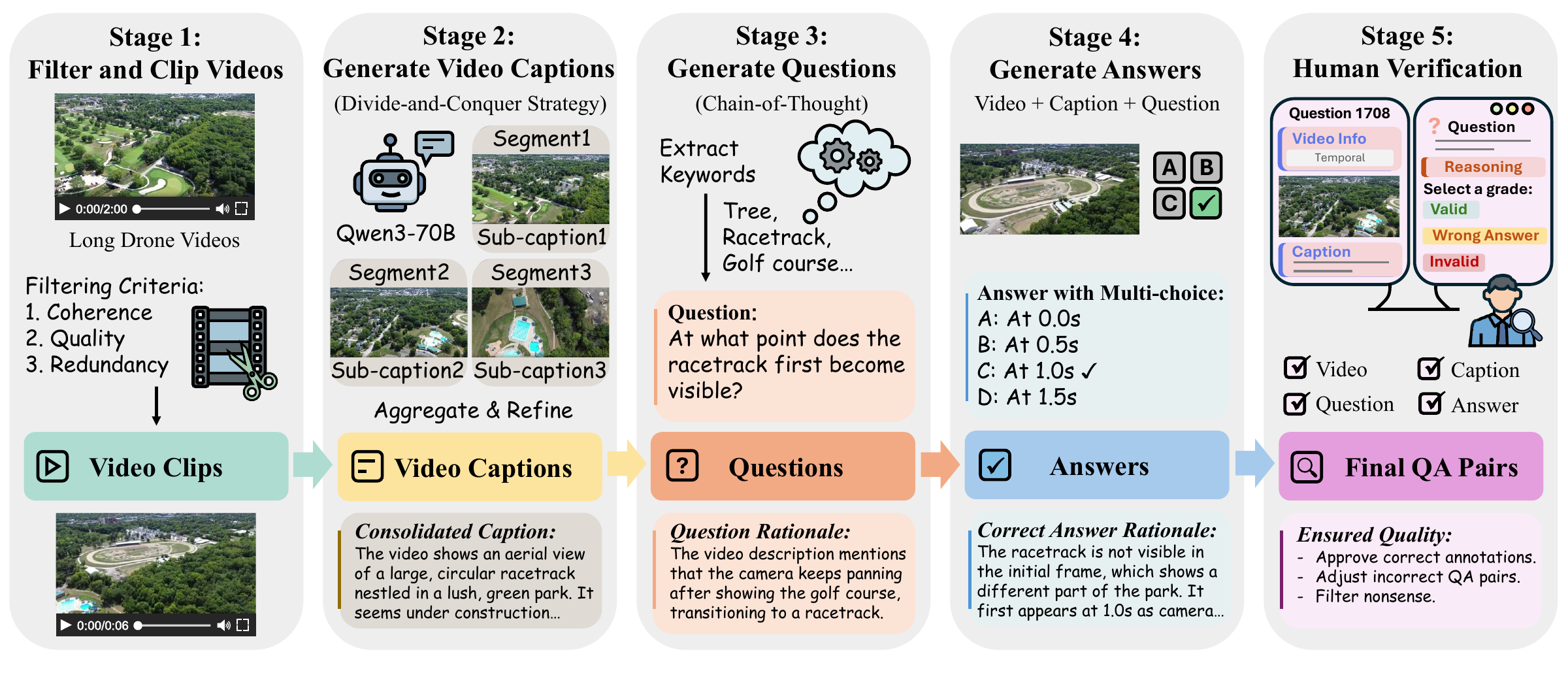}
\vspace{-2\baselineskip}
\caption{\textbf{Overview of the five-stage QA generation workflow.} \textbf{Stage 1}: Long drone video filtering and clipping; \textbf{Stage 2}: Comprehensive video caption generation; \textbf{Stage 3}: Scene-relevant keywords and questions with rationales generation; \textbf{Stage 4}: Multi-choice answers with answer rationales generation; \textbf{Stage 5}: Human verification.}
\vspace{-2\baselineskip}
\label{fig:annotation}
\end{figure*}

\vspace{-0.6\baselineskip}
\subsection{Video Data Curation}\label{subsec:collection}
\vspace{-0.3\baselineskip}
Remote sensing has become an essential field for environmental monitoring \cite{yao2024cracknex}, disaster response \cite{sarkar2023sam}, and traffic analysis \cite{parikh2025roadsocial}. With the rapid advancement of drone technology, high-resolution aerial videos are increasingly available, enabling dynamic observation of complex real-world scenes. \par

For diversity of aerial scenarios, ReVA comprises videos from four sources: 525 from ERA \cite{mou2020era}, 544 from VisDrone \cite{zhu2021detection}, 89 from UAVDT \cite{du2018unmanned}, and 1280 our self-collected videos using a DJI MINI 4 drone. This diverse collection ensures comprehensive coverage of various aerial perspectives and geographic contexts. \par

All videos are resized to 640$\times$360 and trimmed to 15-second clips. Specifically, the ERA subset covers a wide range of activities (e.g., sports and disasters) for scene diversity. The VisDrone subset captures near-ground perspectives at altitudes $<$50 meters, covering urban scenes like city streets and traffic activities. UAVDT comprises long videos for temporal variations. In contrast, our collected videos offer higher altitudes approximately at 100 meters, covering rural scenes. Together, these subsets ensure ReVA covers diverse altitude ranges and scene types from near-ground urban activities to high-altitude rural landscapes, providing comprehensive aerial coverage for thorough evaluation. \par

To prevent scene leakage from training to evaluation, we split the dataset at the original video sequence level rather than clip-level. All clips from the same original video sequence are assigned exclusively to one of the training, validation, or test sets. Therefore, adjacent clips from the same video sequence never appear across different splits, ensuring that training is fully disjoint from testing. \par

\vspace{-0.8\baselineskip}
\subsection{Formulation}
\vspace{-0.5\baselineskip}
We formulate ReVA as a scene-centric video question answering (VideoQA) dataset for remote sensing. Each question-answer pair consists of a video clip $V = \{f_1, f_2, \ldots, f_T\}$ where $f$ is a single frame and a natural language question $Q$. For video-dependent questions, the correct answer requires integrating evidence across multiple frames or over temporal changes, while other questions evaluate scene-level understanding within the video. To facilitate objective evaluation, each question $Q$ is presented in a multiple-choice format with candidate options $C = \{c_1, c_2, \dots, c_n\}$.

\vspace{-0.8\baselineskip}
\subsection{Question-Answer Pair Construction}\label{subsec:anno}
\vspace{-0.5\baselineskip}
Existing RS VisualQA annotation pipelines often scale by enumerating semantic classes with fixed templates (e.g., class existence), yielding repetitive questions. To this end, we propose a five-stage semi-automatic QA generation workflow conditioned on scene context, task objectives, and temporal evidence, producing diverse, scene-centric QA pairs. Fig. \ref{fig:annotation} illustrates the full pipeline. \par 

\noindent \textbf{Video Filtering and Clipping.}
We manually filtered all videos according to three criteria: coherence (narratively inconsistent videos), quality (static or blurry videos), and redundancy (duplicated sequences). The filtered videos are then trimmed to approximately 15-second clips to ensure sufficient scene information for meaningful question generation. \par

\noindent \textbf{Video Caption Generation.}
We adopt a divide-and-conquer strategy to generate comprehensive captions using Qwen3-VL-30B-A3B-Instruct \cite{bai2025qwen3}. Each video is divided into three segments, with Qwen3 generating an individual caption per segment. The full video and segment-level captions are then fed into the MLLM for a consolidated caption capturing global scene context. \par

\noindent \textbf{Question Generation.}
We generate questions via a Chain-of-Thought \cite{wei2022chain} pipeline using caption and video as inputs. The MLLM is prompted to output scene-relevant keywords and generate reasoning-intensive questions with rationales justifying their significance based on keywords. \par

\noindent \textbf{Answer Generation.}
Subsequently, each generated question is fed back into the MLLM alongside the video and consolidated caption to produce a multi-choice answer with its reasoning process. While the rationale is excluded from the final released QA pairs, we empirically find that requiring explicit reasoning encourages deeper inference, yielding more contextually grounded, solid answers.

\begin{figure*}[!t]
\centering
\includegraphics[width=1\linewidth]{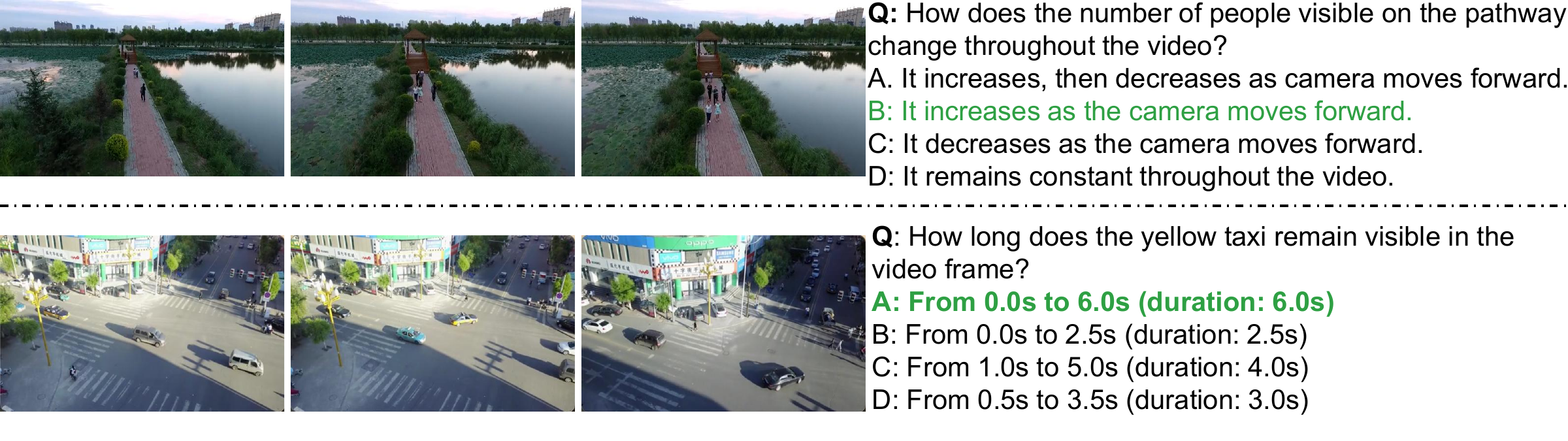}
\vspace{-2\baselineskip}
\caption{\textbf{Examples of ReVA.} Correct answers are marked in \textcolor{softgreen}{green}.}
\label{fig:example}
\vspace{-0.4\baselineskip}
\end{figure*}

\begin{figure*}[!t]
\centering
\vspace{-0.2\baselineskip}
\begin{overpic}[width=0.88\linewidth]{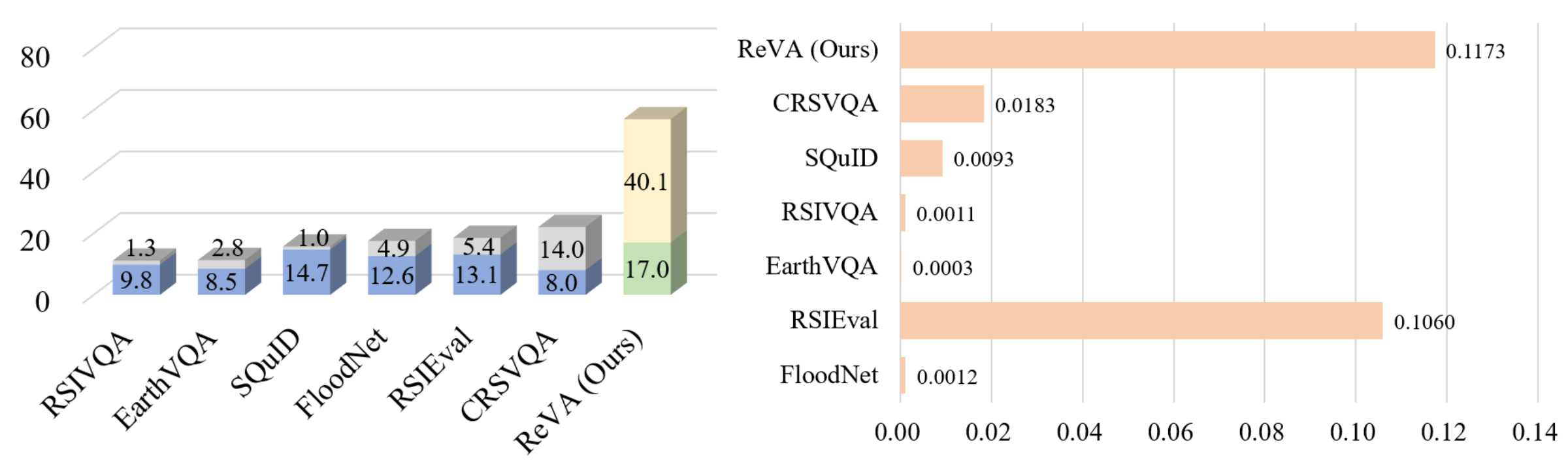}
  \put(10, -1){\small (a) Average token length}
  \put(63, -1){\small (b) 2-gram diversity}
\end{overpic}
\caption{\textbf{QA uniqueness analysis.} ReVA achieves the highest (a) average token length (question on bottom and answer on top) and (b) 2-gram diversities, reflecting superior lexical diversity.}
\label{fig:token_length}
\vspace{-1.5\baselineskip}
\end{figure*}

\noindent \textbf{Human Verification.}
To reduce LLM hallucinations, all automatically generated QA pairs undergo a multi-stage human verification process. In the first stage, eight trained reviewers inspect assigned QA pairs for correctness of the answer and question-answer-option quality. Each QA pair is reviewed by two reviewers. Reviewers either (1) accept the QA pair, (2) make minor corrections, (3) reject the QA, or (4) flag it for further review. In the second stage, two additional expert reviewers examine flagged samples and adjudicate ambiguous cases. QA pairs that cannot be reliably grounded in the video are removed. More details are in the Appendix. Fig. \ref{fig:example} shows two examples of our data. \par

\begin{figure*}[t]
\centering
\includegraphics[width=0.9\linewidth]{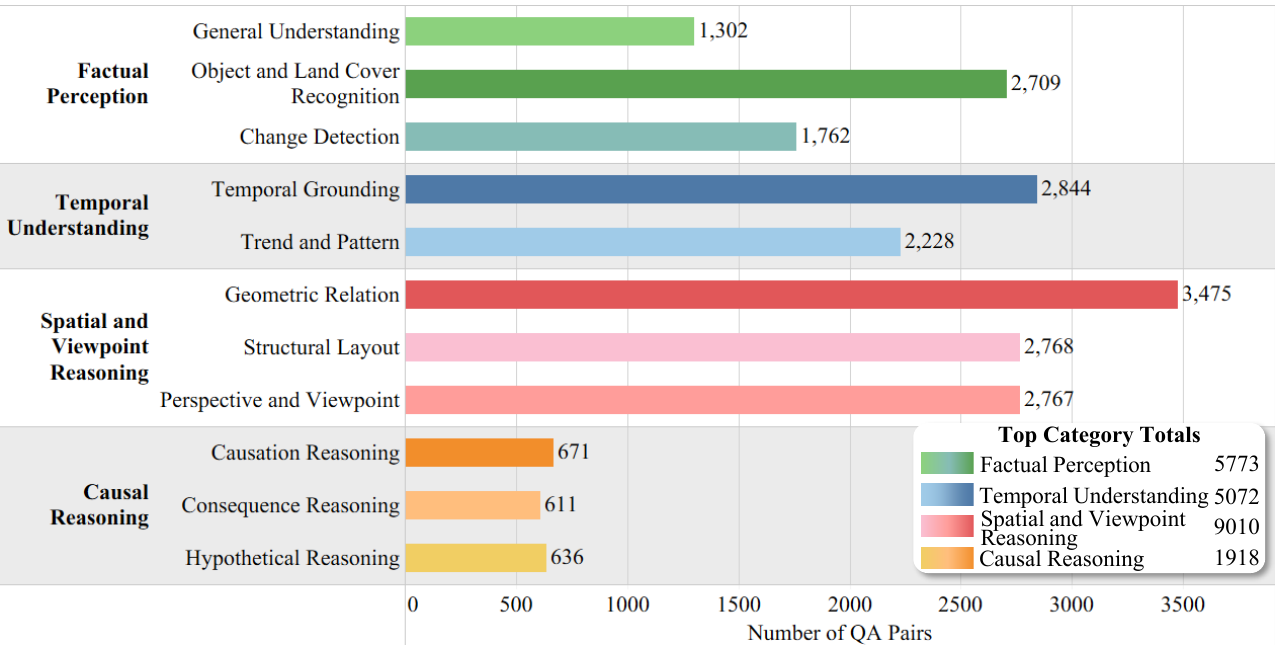}
\vspace{-1.2\baselineskip}
\captionof{figure}{\textbf{Question distribution by categories and tasks.}}
\label{fig:distribution}
\vspace{-1.6\baselineskip}
\end{figure*}

\vspace{-0.6\baselineskip}
\subsection{Statistics and Analysis} \label{subsec:stats}
\vspace{-0.3\baselineskip}

\noindent \textbf{Geographical Distribution.}
ReVA spans 18 cities, 14 from China (VisDrone) and 4 from the United States (rural areas) spanning Illinois, Pennsylvania, and New York (our collected videos):
\vspace{-1mm}
\begin{itemize}
    \item \textbf{China:} Tianjin, Hong Kong, Daqing, Ganzhou, Guangzhou, Jinchang, Liuzhou, Nanjing, Shaoxing, Shenyang, Nanyang, Zhangjiakou, Suzhou, and Xuzhou.
    \item \textbf{United States:} Champaign (IL), Philadelphia (PA), Bethlehem (PA), and New York (NY).
\end{itemize}
\vspace{-1mm}

The videos comprise multiple viewpoints (e.g., top-down and low-angle view) and cover diverse scene elements such as roads and farmlands, recorded under varying weather conditions. Geographic metadata for ERA is unavailable, as videos were collected from public internet sources. \par

\begin{wraptable}{r}{0.35\linewidth}
\vspace{-1.5\baselineskip}
\small
\caption{\textbf{ReVA Statistics.}}
\vspace{-0.8\baselineskip}
\label{tab:ReVA_stats}
\centering
\resizebox{0.9\linewidth}{!}{%
\begin{tabular}{l|cc}
\toprule
 Split & Video & QA pairs \\
\midrule
Train & 1,045 & 15,773 \\
Val & 379 & 2,000 \\
Test & 1,014 & 4,000 \\
\midrule
Total & 2,438 & 21,773 \\
\bottomrule
\end{tabular}}
\vspace{-1.5\baselineskip}
\end{wraptable}

\noindent \textbf{Dataset Statistics.} As summarized in Table \ref{tab:ReVA_stats}, ReVA contains 2,438 videos, with 15,773, 2,000, and 4,000 QA pairs in the training, validation, and test sets, respectively. The corresponding split-level video counts are 1,045, 379, and 1,014. The training videos are kept disjoint from both evaluation subsets. As shown in Fig. \ref{fig:token_length}, ReVA achieves the highest average token length and 2-gram diversity, demonstrating greater linguistic diversity and scene-centric QA pairs. \par

\noindent \textbf{Question Taxonomy.}
We organize ReVA questions into 4 major categories with 11 QA tasks as shown in Fig. \ref{fig:distribution}: \textbf{(1) Factual Perception} focuses on observable visual content in the scene, comprising 5,773 QA pairs: (a) \emph{General Understanding} with overall scene-level perception (1,302); (b) \emph{Object and Land Cover Recognition} covering brief questions like existence, counting, and classification (2,709); and (c) \emph{Change Detection} with differences across frames (1,762). \textbf{(2) Temporal Understanding} evaluates capabilities of reasoning over time and aggregating cross-frame information, comprising 5,072 samples: (a) \emph{Temporal Grounding} with precise timestamp localization (2,844); and (b) \emph{Trend and Pattern} demanding temporal reasoning of dynamic processes (2,228). \textbf{(3) Spatial and Viewpoint Reasoning} measures spatial relational reasoning and camera-scene interaction, comprising 9,010 samples: (a) \emph{Geometric Relation} with spatial relationships between objects (3,475); (b) \emph{Structural Layout} for spatial arrangement and network structure patterns such as roads and rivers (2,768); and (c) \emph{Perspective and Viewpoint} covering camera states and motion understanding (2,767). \textbf{(4) Causal Reasoning} assesses higher-level logical inference grounded in observable evidence, comprising 1,918 samples: (a) \emph{Causation Reasoning} identifying cause-effect relationships (671); (b) \emph{Consequence Reasoning} predicting plausible outcomes from observed trends (611); and (c) \emph{Hypothetical Reasoning} for counterfactual reasoning against direct observation (636). This hierarchical taxonomy enables temporal aggregation and spatial reasoning rather than low-level perception only, establishing a comprehensive benchmark for scene-centric, reasoning-intensive remote sensing video understanding. \par

\vspace{-1\baselineskip}
\section{ReMoSense: Motion-Aware Remote Sensing VideoQA} \label{sec:method}
\vspace{-0.6\baselineskip}
\subsection{Motivation}
\vspace{-0.3\baselineskip}
Remote sensing drone videos exhibit motion patterns distinct from natural videos: dominant camera ego-motion from viewpoint and altitude changes coexists with small, subtle object dynamics. Generic video models typically learn motion implicitly, making it difficult to separate camera-induced motion from true scene changes, especially under limited remote sensing supervision. This ambiguity disrupts cross-frame correspondence and degrades viewpoint and long-range temporal understanding. \par

To address these limitations, we propose ReMoSense, a motion-aware framework for remote sensing video understanding that explicitly disentangles inter-frame correspondence motion from object temporal dynamics. By modeling dual motion as explicit, structured inputs rather than implicit features, ReMoSense directly mitigates the motion ambiguity, improving spatiotemporal reasoning. \par

\begin{figure*}[!t]
\centering
\includegraphics[width=0.8\linewidth]{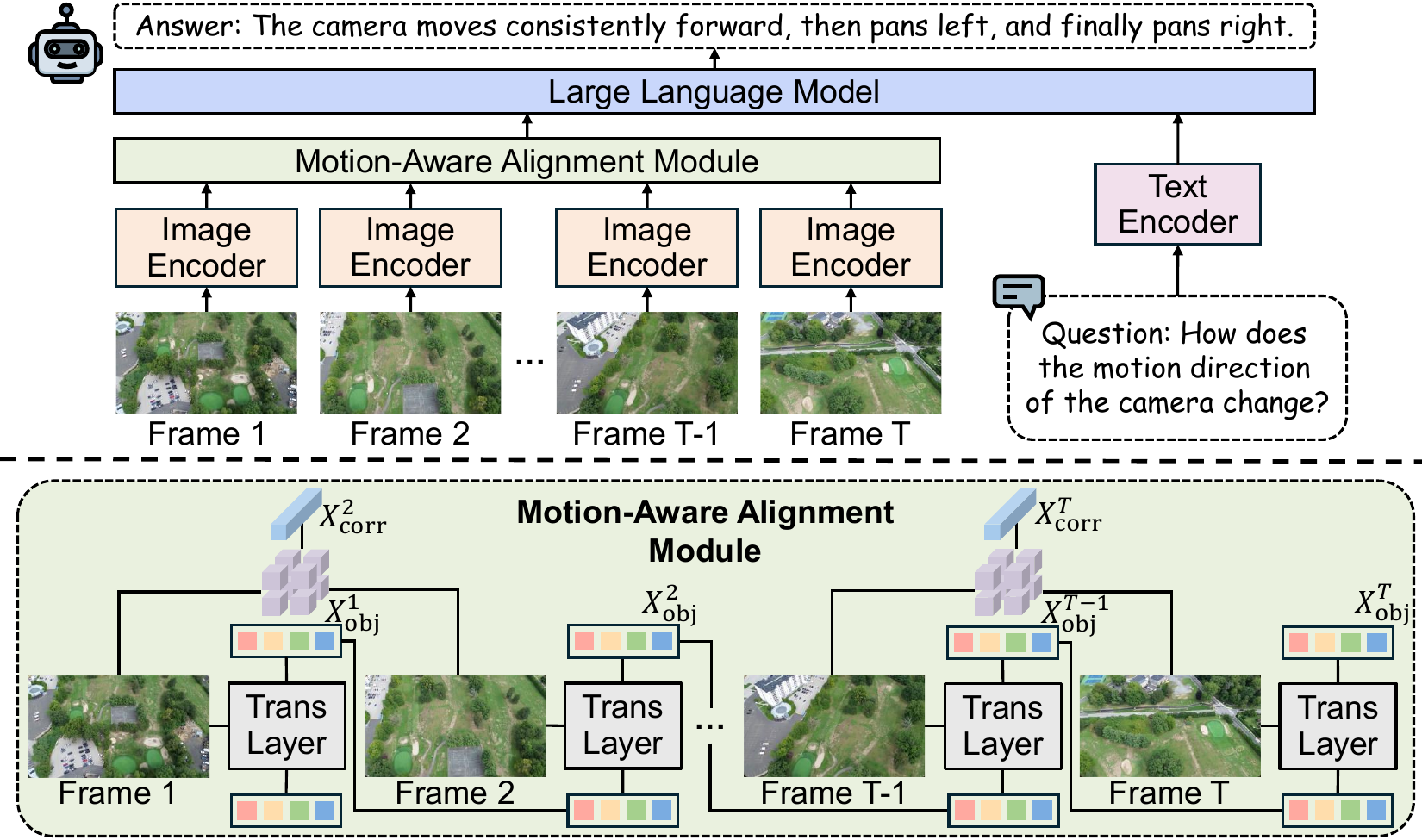}
\vspace{-0.4\baselineskip}
\caption{\textbf{Overview of ReMoSense Framework and Motion-Aware Alignment Module.} The Motion-Aware Alignment Module captures inter-frame correspondence for global motion alignment. It then captures salient object motion iteratively for long-range temporal consistency. }
\label{fig:model}
\vspace{-1.4\baselineskip}
\end{figure*}

\vspace{-0.6\baselineskip}
\subsection{Architecture Overview}
\vspace{-0.3\baselineskip}
Fig. \ref{fig:model} provides an overview of our proposed ReMoSense. Our model follows a standard next-token prediction paradigm and consists of 4 components: an image encoder \cite{dosovitskiy2020image}, a text encoder, a Motion-Aware Alignment Module, and an LLM decoder. We first encode each input frame independently to extract patch-level visual tokens $X_\mathrm{img}$. The alignment module then computes cost volumes between consecutive frames to derive global correspondence tokens $X_\mathrm{corr}$ for cross-frame correlation. In parallel, we initialize learnable object motion tokens $X_\mathrm{obj}$ at $t=1$ and propagate them. For each subsequent frame $t=2,\ldots,T$, previous object tokens $X_\mathrm{obj}^{t-1}$ are updated $X_{\mathrm{obj}}^{t}$ using the current visual tokens to iteratively refine object dynamics. Finally, we concatenate visual tokens and two motion tokens, generating a motion-enhanced visual representation $X^{1:T}=[X_\mathrm{img}^{1:T}; X_\mathrm{corr}^{1:T}; X_\mathrm{obj}^{1:T}]$ with the encoded question and feed the sequence $X$ to decoder for final answer. \par

\vspace{-0.6\baselineskip}
\subsection{Dual Motion-Aware Alignment}
\vspace{-0.3\baselineskip}
ReMoSense introduces motion-aware alignment by modeling global correspondence tokens that encode general inter-frame visual changes via cost volumes, and object motion tokens that auto-regressively update temporal dynamics of salient objects across frames for long-range dependency. These two key components form our Dual Motion-Aware Alignment Module, which injects explicit geometric motion, enhancing spatiotemporal coherence for long-range reasoning. \par

\noindent \textbf{Global Inter-frame Correspondence.}
Given visual tokens $\{X_\mathrm{img}^{n}, X_\mathrm{img}^{n+1}\} \in\mathbb{R}^{H\times W\times C}$ for two consecutive frames, we construct cost volume maps $\mathcal{C}^{n+1} \in\mathbb{R}^{HW/16 \times HW/16}$ by computing pairwise similarities between their downsampled patch tokens ($X_\mathrm{img}^{n}$ and $X_\mathrm{img}^{n+1}$ to $H/4 \times W/4\times C$ resolution):
\begin{equation}
\mathcal{C}^{n+1}(i,j)
= \frac{X_\mathrm{img}^n(i)'^\top X_\mathrm{img}^{n+1}(j)'}
{\left\|X_\mathrm{img}^n(i)'\right\|_2 \, \left\|X_\mathrm{img}^{n+1}(j)'\right\|_2},
\quad i,j \in \{1,\dots,HW/16\},
\end{equation}
where $X_\mathrm{img}'$ are downsampled visual tokens. We then expand each scalar value through an MLP, yielding a channel-expanded correspondence representation $\mathcal{C}^{n+1\,\prime}\in\mathbb{R}^{HW/16 \times HW/16 \times C}$. Finally, we apply global pooling over the two spatial dimensions to obtain a compact global correspondence token $X_\mathrm{corr}^{n+1} \in\mathbb{R}^{1 \times C}$ and concatenate it to $X_\mathrm{img}^{n+1}$. This representation summarizes aggregate inter-frame correspondence patterns associated with global visual changes, such as those induced by camera and viewpoint movement, and provides compact motion-aware cue for cross-frame reasoning. \par

\noindent \textbf{Iterative Object Motion Modeling.}
To capture temporally consistent object dynamics in an end-to-end framework, we introduce learnable object motion tokens $X_{\mathrm{obj}} \in \mathbb{R}^{K \times C}$ propagated auto-regressively across frames. At $t=1$, we initialize $X_{\mathrm{obj}}^{t=1}$ as learnable parameters. For each subsequent frame $t=n$, previous queries $X_{\mathrm{obj}}^{n-1}$ are updated via Transformer layers, injecting frame-specific visual features $X_{\mathrm{img}}^{n}$ while preserving temporal states. The output $X_{\mathrm{obj}}^n$ is a refined object-centric representation that follow subtle motions and appearance changes across frames. We repeat this refinement process over $t=1:T$, yielding a sequence of temporally aligned object tokens that encode motion and appearance changes across frames. These tokens are concatenated with their corresponding frame's visual tokens, as input to the LLM decoder, augmenting each frame with an object-centric association. This recurrent design provides a lightweight mechanism for maintaining object consistency over long horizons, enhancing long-range dependency in dynamic aerial scenes. \par

%% file: sec/4_result.tex
\begin{table*}[!t]
\centering
\caption{\textbf{Accuracy (\%) on ReVA test set.} The best-performing results are presented in \textbf{bold}, while the second-best results are \underline{underlined}.}
\vspace{-0.6\baselineskip}
\resizebox{\textwidth}{!}{
\begin{tabular}{l|>{\centering\arraybackslash}p{0.68cm}|ccccccccccc|>{\centering\arraybackslash}p{1cm}}
\toprule
\multirow{3}{*}{Model}
& \multirow{3}{*}{Size}
& \multicolumn{3}{c}{Factual}
& \multicolumn{2}{c}{Temporal}
& \multicolumn{3}{c}{Spatial}
& \multicolumn{3}{c|}{Causal}
& \multirow{3}{*}{Overall} \\

& & \multicolumn{3}{c}{Perception}
& \multicolumn{2}{c}{Understanding}
& \multicolumn{3}{c}{Reasoning}
& \multicolumn{3}{c|}{Reasoning} & \\

\cmidrule(lr){3-5}\cmidrule(lr){6-7}\cmidrule(lr){8-10}\cmidrule(lr){11-13} 
& & GU & OR & CD & TG & TP & GR & SL & PV & CA & CO & HR &  \\
\midrule

\rowcolor{lightdark} \multicolumn{14}{c}{\textbf{Based on Proprietary MLLMs}} \\
LLoVi & -
& 86.67 & 65.15 & 55.00
& 49.06 & 68.33
& 63.75 & 76.76 & 55.21
& 88.89 & 90.00 & 85.62
& 64.73 \\

VideoTree & -
& 75.00 & 46.67 & 49.60
& 45.63 & 63.33
& 48.50 & 64.12 & 50.63
& 88.89 & 83.00 & 86.88
& 56.20 \\

VideoAgent & -
& 74.44 & 43.33 & 47.20
& 39.69 & 59.44
& 46.25 & 62.94 & 52.71
& 85.56 & 79.00 & 85.00
& 53.62 \\

\rowcolor{lightbrown}
\multicolumn{14}{c}{\textbf{Open-source MLLMs Trained on Our 22K ReVA Video Data}} \\

LLaVA-NeXT-Video & 7B
& 78.33 & 55.76 & 46.20
& 39.06 & 51.67
& 50.75 & 62.35 & 39.79
& 83.89 & 82.00 & 65.00
& 52.98 \\

VideoChat-Flash & 7B
& \best{96.67} & \secondbest{77.42} & 66.60
& 60.47 & \secondbest{74.44}
& 70.50 & 77.06 & 61.88
& 91.67 & 92.00 & 83.12
& 72.60 \\

Video-LLaVA & 7B
& 62.78 & 49.55 & 44.40
& 64.22 & 59.44
& 64.00 & 56.47 & 61.46
& 86.67 & 81.00 & 60.62
& 59.10 \\

VideoLLaMA2 & 7B
& 92.22 & 71.51 & \best{71.20}
& \secondbest{73.70} & 73.56
& \secondbest{70.75} & \secondbest{79.64} & 76.68
& 92.22 & \best{98.00} & \secondbest{85.62}
& \secondbest{76.33} \\

BIMBA & 7B
& 94.44 & 72.42 & 64.40
& 64.06 & 63.89
& 69.00 & 78.24 & \secondbest{80.83}
& \secondbest{94.44} & 96.00 & 83.75
& 73.50 \\

\textbf{ReMoSense (Ours)} & 7B
& \secondbest{96.12} & \best{78.03} & \secondbest{70.85}
& \best{76.21} & \best{76.40}
& \best{74.25} & \best{83.27} & \best{83.74}
& \best{95.32} & \secondbest{97.65} & \best{90.74}
& \best{80.04} \\

\rowcolor{lightblue}
\multicolumn{14}{c}{\textbf{Open-source MLLMs Trained on Larger Scale Instruction LLaVA-Video-178K Data}} \\

NVILA & 3B
& 88.89 & 57.88 & 27.20
& 45.62 & 35.00
& 42.00 & 64.71 & 50.42
& 43.33 & 62.00 & 60.00
& 49.05 \\

VideoLLaMA2 & 7B
& 78.89 & 57.12 & 56.00
& 47.50 & 68.89
& 43.00 & 61.18 & 41.88
& 90.56 & 88.00 & 84.38
& 57.95 \\

LLaVA-NeXT-Video & 7B
& 95.00 & 75.61 & 59.00
& 40.47 & 66.94
& 69.00 & 72.06 & 60.21
& 88.89 & 87.00 & 82.50
& 66.35 \\

VideoChat-Flash & 7B
& 94.44 & 74.70 & \secondbest{63.20}
& 56.72 & \secondbest{72.22}
& 69.50 & 70.00 & 55.00
& 92.78 & 93.00 & 83.75
& 69.40 \\

Video-LLaVA & 7B
& 88.89 & 65.76 & 57.20
& \secondbest{59.06} & 56.11
& 68.00 & \secondbest{77.06} & 54.17
& 91.11 & 90.00 & 82.50
& 66.00 \\

BIMBA & 7B
& \best{97.78} & 75.15 & 59.60
& \best{65.62} & 65.56
& 61.50 & 73.53 & \secondbest{80.83}
& 94.44 & \secondbest{96.00} & 86.25
& \secondbest{72.85} \\

MovieChat-OneVision & 7B
& \secondbest{95.56} & 73.64 & 62.00
& 55.31 & 63.33
& 60.00 & 62.35 & 35.83
& \secondbest{94.78} & \secondbest{96.00} & 83.75
& 64.37 \\

VideoMind & 7B
& 92.22 & 70.45 & 60.40
& 32.66 & 63.33
& 65.50 & 67.06 & 53.75
& 90.56 & 87.00 & 80.00
& 62.40 \\

VITAL & 7B
& 85.56 & 68.03 & 57.60
& 38.12 & 68.89
& 65.50 & 69.41 & 67.29
& 87.22 & 84.00 & 83.75
& 64.48 \\

Tarsier2 & 7B
& 93.89 & 74.09 & 62.20
& 41.41 & 63.61
& 59.25 & 67.94 & 58.96
& 90.00 & 82.00 & 80.00
& 64.65 \\

AVATAR & 7B
& 34.44 & 43.94 & 49.40
& 37.81 & 56.39
& 62.25 & 54.12 & 41.04
& 85.00 & 87.00 & 78.12
& 50.98 \\

InternVL3 & 7B
& 94.44 & 71.36 & 63.00
& 40.00 & 68.06
& 70.75 & 70.88 & 59.79
& 90.00 & 85.00 & 85.00
& 66.28 \\

NVILA & 8B
& 93.33 & \best{77.88} & 55.20
& 36.25 & 70.00
& 67.50 & 61.76 & 68.33
& 90.00 & 92.00 & 82.50
& 65.90 \\



Gemma3 & 27B
& 92.22 & \secondbest{76.67} & 57.40 
& 42.03 & 65.28
& 69.25 & 75.00 & 58.96
& 89.44 & 90.00 & \secondbest{87.50}
& 66.72 \\

Nemotron3-Nano-Omni & 30B
& 93.89 & 72.73 & 60.80
& 38.91 & 69.72
& \secondbest{71.75} & 72.35 & 63.12
& 90.56 & 95.00 & 83.12
& 67.00 \\


\textbf{ReMoSense (Ours)} & 7B
& \best{97.78} & 74.85 & \best{67.20}
& 58.44 & \best{75.00}
& \best{78.00} & \best{85.29} & \best{81.25}
& \best{96.67} & \best{98.00} & \best{90.00}
& \best{76.45} \\

\bottomrule
\end{tabular}
}
\begin{tablenotes}[flushleft]
\scriptsize
\item * GU: General Understanding, OR: Object and Land Cover Recognition, CD: Change Detection, TG: Temporal Grounding, TP: Trend and Pattern, GR: Geometric Relation, SL: Structural Layout, PV: Perspective and Viewpoint, CA: Causation Reasoning, CO: Consequence Reasoning, and HR: Hypothetical Reasoning.
\end{tablenotes}
\vspace{-1.6\baselineskip}
\label{tab:reva_test}
\end{table*}

\vspace{-1\baselineskip}
\section{Experiments} \label{sec:expt}
\vspace{-0.6\baselineskip}
\subsection{Evaluation Details} \label{sec:evaldetails}
\vspace{-0.3\baselineskip}
\noindent \textbf{Experimental Setting.} 
To thoroughly evaluate video reasoning capabilities of MLLMs on ReVA, we selected: (1) Proprietary MLLMs, e.g., LLoVi \cite{zhang2024simple}, VideoAgent \cite{wang2024videoagent}, and VideoTree \cite{wang2025videotree}; (2) Open-source MLLMs fine-tuned on ReVA training set, e.g., VideoChat-Flash \cite{li2024videochat}, LLaVA-NeXT-Video \cite{zhang2024llavanextvideo}, VideoLLaVA \cite{lin2024video}, VideoLLaMA2 \cite{cheng2024videollama}, BIMBA \cite{islam2025bimba}; (3) Open-source MLLMs from 3B to 30B trained on large instruction video data, e.g., MovieChat-OneVision \cite{song2025moviechat}, InternVL3 \cite{zhu2025internvl3}, NVILA \cite{liu2025nvila}, VideoMind \cite{liu2025videomind}, VITAL \cite{zhang2025thinking}, Tarsier2 \cite{yuan2025tarsier2}. Note that for (3), we test baselines pre-trained on the LLaVA-Video-178K dataset \cite{zhang2024llava} following common practice \cite{islam2025bimba}. \par

To ensure fair comparison, all baselines trained on ReVA video data are fine-tuned for 24K steps with a global batch size of 1. All other training details including frame sampling rate and other hyper-parameters follow their respective original settings. We set maximum output token length to 512 following common practice. For our model training, we use AdamW optimizer \cite{loshchilov2017decoupled} with a learning rate of 2e-4, cosine decay schedule, and warmup ratio of 0.03 under DeepSpeed ZeRO-1 framework \cite{rasley2020deepspeed}. All training and evaluation are conducted on 2 NVIDIA 80GB A100 GPUs. \par

\noindent \textbf{Implementation Details.} For ReMoSense, we set number of object motion token to $K=16$ and they are randomly initialized. We downsample visual tokens from $64\times 64$ to $16\times 16$ when computing cost volumes, which reduces the overhead to only 10\%. We use standard cross-entropy loss
for text generation following common practice \cite{islam2025bimba}. \par

\vspace{-0.6\baselineskip}
\subsection{Quantitative Results}
\vspace{-0.3\baselineskip}
We evaluate baseline model performance on ReVA test set in Table \ref{tab:reva_test}. When fine-tuned on ReVA, our proposed method, ReMoSense, achieves 80.04\% overall accuracy, outperforming the previous best fine-tuned baseline VideoLLaMA2 by 3.7\%. Compared to our baseline model with the same backbone, BIMBA, the performance improves 6.5\%. When trained on large-scale data, ReMoSense achieves 76.45\% overall accuracy, outperforming the previous best model, BIMBA, by 3.6\%. Among all categories, our model demonstrates the most improvements on Geometric Relation (+6.3\%) and Structural Layout (+8.2\%). \par

\begin{table*}[!t]
\centering
\caption{\textbf{Ablation study for various design choices.} The best-performing results are presented in \textbf{bold}, while the second-best results are \underline{underlined}.}
\vspace{-0.6\baselineskip}
\resizebox{\linewidth}{!}{%
  \begin{tabular}{c|cc|cccccccccccc}
  \toprule
      \multirow{3}{*}{Variant}
    & \multirow{3}{*}{Global}
    & \multirow{3}{*}{Object}
    & \multicolumn{3}{c}{Factual}
    & \multicolumn{2}{c}{Temporal}
    & \multicolumn{3}{c}{Spatial}
    & \multicolumn{3}{c|}{Causal}
    & \multirow{3}{*}{Overall} \\
    & & & \multicolumn{3}{c}{Perception}
    & \multicolumn{2}{c}{Understanding}
    & \multicolumn{3}{c}{Reasoning}
    & \multicolumn{3}{c|}{Reasoning} & \\
    \cmidrule(lr){4-6}\cmidrule(lr){7-8}\cmidrule(lr){9-11}\cmidrule(lr){12-14} 
    & & & GU & OR & CD & TG & TP & GR & SL & PV & CA & CO & HR &  \\
  \midrule
I &  &   
& 94.44 & 72.42 & 64.40
& 64.06 & 63.89
& 69.00 & 78.24 & 80.83
& \secondbest{94.44} & \secondbest{96.00} & 83.75
& 73.50 \\

II & \checkmark &
& 95.00 & 74.85 & \secondbest{69.80}
& 65.62 & 65.28
& 70.50 & 78.24 & 81.25
& 92.78 & 93.00 & 87.50
& 75.18 \\

III & & \checkmark
& \secondbest{95.56} & \secondbest{76.67} & 67.20
& \secondbest{67.18} & \secondbest{65.66}
& \secondbest{71.50} & \secondbest{80.94} & \secondbest{81.50}
& \secondbest{94.44} & \secondbest{96.00} & \best{91.25}
& \secondbest{76.12} \\

IV & \checkmark & \checkmark
& \best{96.12} & \best{78.03} & \best{70.85}
& \best{76.21} & \best{76.40}
& \best{74.25} & \best{83.27} & \best{83.74}
& \best{95.32} & \best{97.65} & \secondbest{90.74}
& \best{80.04} \\
\bottomrule
\end{tabular}
}
\label{tab:ablation}
\end{table*}

\begin{figure*}[t]
\centering
\vspace{-0.6\baselineskip}
\begin{overpic}[width=\linewidth]{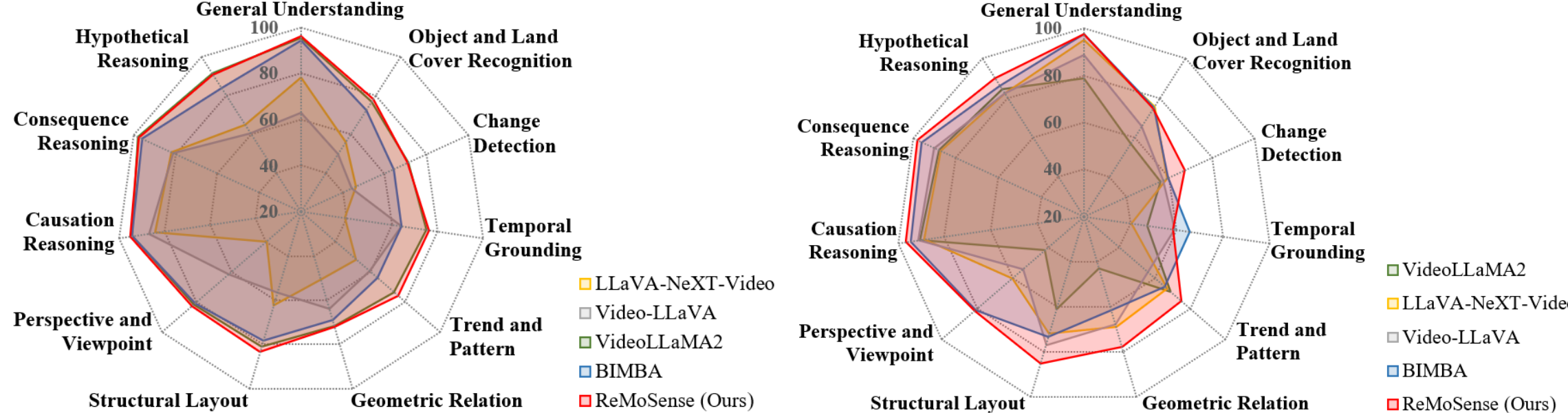}
  \put(2, -3){\small (a) Open-source MLLMs trained on ReVA}
  \put(48, -3){\small (b) Open-source MLLMs trained on large-scale data}
\end{overpic}
\vspace{0.2\baselineskip}
\caption{\textbf{Per-task comparisons with SOTA methods on ReVA test set.}}
\vspace{-1.8\baselineskip}
\label{fig:radar}
\end{figure*}

\vspace{-0.6\baselineskip}
\subsection{Ablation Study} \label{sec:ab}
\vspace{-0.3\baselineskip}
Table \ref{tab:ablation} presents ablation studies of ReMoSense when trained on ReVA data with the same Qwen2.5-VL backbone: (I) baseline; (II) ReMoSense without object motion tokens; (III) ReMoSense without global correspondence tokens; (IV) full model. Removing global correspondence tokens leads to an average drop of 3.9\%. Among all tasks, Temporal Grounding decreases by 9.0\%, confirming its role in capturing cross-frame dynamics. Removing object motion tokens causes 4.9\% degradation. The proposed global motion complements the object motion by providing scene-level temporal context. The full model achieves the best performance, showing the complementary nature of both tokens. \par

\vspace{-0.6\baselineskip}
\subsection{Discussion}
\vspace{-0.3\baselineskip}
Our analysis reveals three key limitations of current MLLMs on remote sensing video understanding. (1) Existing models exhibit a systematic bias of spatial priors: they tend to output descriptions such as ``left-top corner'' when they are uncertain, suggesting pretrained spatial heuristics over true geometric reasoning. (2) Their failures are particularly evident in temporal grounding and viewpoint reasoning, where models confuse activity order, over-rely on visually salient key frames, and fail to disentangle camera ego-motion from object motion. It indicates that video MLLMs lack explicit mechanisms for modeling inter-frame continuity and orientation. (3) Results from ReVA fine-tuning show that domain-aligned training is more effective than simply scaling model size, especially for spatiotemporal tasks. These findings show that remote sensing videoQA requires benchmarks and models that go beyond generic video-language pretraining. More details are in the Appendix. \par

\vspace{-0.5\baselineskip}
\section{Conclusion} \label{sec:conclusion}
\vspace{-0.5\baselineskip}
In this paper, we present ReVA, a fully real-world remote sensing VideoQA dataset designed for scene-centric understanding, comprising 22K QA pairs across 2,438 videos from 18 cities and 17K unique questions spanning 11 tasks. We further develop a five-stage semi-automatic QA generation pipeline that ensures scalability and facilitates future dataset construction. We thoroughly evaluate 23 proprietary and open-source Video LLMs on ReVA, revealing fundamental limitations in temporal grounding and viewpoint reasoning under aerial perspectives. To address these gaps, we propose ReMoSense, a motion-aware framework that explicitly models global correspondence motion for robust cross-frame alignment and object motion for long-range temporal reasoning. Together, ReVA and ReMoSense advance the boundary of remote sensing video understanding, offering a promising direction and a strong baseline for future remote sensing reasoning research. \par

%% file: sec/X_suppl.tex
\newpage
\appendix
\section*{Appendix Overview}
In this Appendix, we provide additional details of the paper, including discussion \ref{sec:discussion}, additional implementation details (Section \ref{sec:adimplement}), additional results (Section \ref{sec:addresult}), additional dataset details (Section \ref{sec:addataset}), question type details (Section \ref{sec:question_type_detail}), future work (Section \ref{sec:future}), privacy policy (Section \ref{sec:privacy}), prompt details (Section \ref{sec:prompt_detail}), and examples by categories (Section \ref{sec:example}). \par

\section{Discussion} \label{sec:discussion}
\noindent \textbf{Observation 1: Spatial Reasoning Bias.}
Beyond object localization, we observe a systematic bias showing spurious spatial priors. When highly uncertain, several models tend to have biased, fixed spatial descriptions of "left-top corner" regardless of the true spatial configuration. It suggests that spatial heuristics acquired during pretraining might override implicit geometric reasoning. \par

This qualitative finding demonstrates that spatial reasoning in ReVA requires structured design for geometric modeling. Errors typically arise not from inaccurate object recognition, but from incorrect reasoning about orientation, viewpoint, and relative configuration due to the spatial bias. \par

\noindent \textbf{Observation 2: Temporal and Viewpoint Failures.} 
We further observe systematic failures of SOTA models on temporal grounding and viewpoint-reasoning tasks, which are identified as two principal failure modes: (1) Temporal order confusion: existing models tend to confuse event ordering (e.g., before/after). This is because they over-emphasize visually salient key frames while neglecting inter-frame continuity. This reflects a reliance on frame-level semantic salience rather than explicit sequential modeling. (2) Viewpoint mismatch: these models fail to distinguish camera-induced ego-motion from scene dynamics, treating global background shifts as independent transitions. Consequently, they struggle with spatial relationships such as orientation and directions over correctly localized objects. For example, they predict the motion of a car with left-to-right motion as an opposite trajectory when correctly detected. \par

These observations are quantitatively verified in Table \ref{tab:reva_results}, where spatiotemporal tasks are generally the weakest sub-tasks. The fact that pre-training on large-scale video data fails to solve these limitations reveals the significance of robust temporal and viewpoint localization on remote sensing, demonstrating the necessity of both ReVA and ReMoSense. \par


\noindent \textbf{Observation 3: Dataset-Level Implications.} Model results trained on ReVA reveal that domain-aligned fine-tuning is more effective than scaling model size. Fine-tuning on ReVA yields substantial gains on spatiotemporal sub-tasks (e.g., Temporal Grounding and Viewpoint and Perspective) by 10–20\%. Tasks relying primarily on language-level reasoning (although visually grounded), such as Causal Reasoning, show limited incremental gains, suggesting pre-training saturation. This asymmetry indicates that domain-aligned fine-tuning benefits explicit temporal localization and spatial structure modeling, capabilities that general domain video pre-training at scale fails to achieve. \par

\section{Additional Implementation Details} \label{sec:adimplement}
\noindent \textbf{Model details.} We adopt Qwen2.5-VL-7B-Instruct\cite{bai2023qwen} as our backbone, which consists of a Qwen2.5 visual encoder and a Qwen2.5 text encoder (using re-engineered ViT \cite{dosovitskiy2020image}) connected via a patch merger. We present a two-stage training pipeline. In the first stage, all backbone parameters are frozen and only the proposed Motion-Aware Alignment Module is optimized. In the second stage, we apply LoRA \cite{hu2022lora} tuning method to further improve the performance, with the rank of LoRA is 16 and the alpha is 32. \par

\noindent \textbf{Training details.} The parameters for the AdamW optimizer \cite{loshchilov2017decoupled} are $\beta_1=0.9$, $\beta_2=0.999$, $\epsilon=10^{-8}$. The gradient clipping is set to be 1.0. During training, the batch size per device is set to be 1. For large open-source MLLMs (30B), we use their released checkpoints for evaluation only; no additional fine-tuning is performed. \par
 
\noindent \textbf{Data preprocessing.} We uniformly sample 32 frames from each video, following the common practice in video understanding \cite{islam2025bimba}. All frames are resized to $640\times360$. Our configuration provides sufficient spatial and temporal detail while maintaining low computational cost during training and inference. \par

\noindent \textbf{Evaluation Metrics.} Since all QA pairs in ReVA are multiple-choice, we report top-1 accuracy (the percentage of question answered correctly). \par

\renewcommand{\arraystretch}{1.05}
\begin{table*}[t]
\centering
\caption{\textbf{Accuracy (\%) on ReVA validation set.} For each block, the best-performing results are presented in \textbf{bold}, while the second-best results are \underline{underlined}.}
\resizebox{\textwidth}{!}{
\begin{tabular}{l|>{\centering\arraybackslash}p{0.68cm}|ccccccccccc|>{\centering\arraybackslash}p{1cm}}
\toprule
\multirow{3}{*}{Model}
& \multirow{3}{*}{Size}
& \multicolumn{3}{c}{Factual}
& \multicolumn{2}{c}{Temporal}
& \multicolumn{3}{c}{Spatial}
& \multicolumn{3}{c|}{Causal}
& \multirow{3}{*}{Overall} \\

& & \multicolumn{3}{c}{Perception}
& \multicolumn{2}{c}{Understanding}
& \multicolumn{3}{c}{Reasoning}
& \multicolumn{3}{c|}{Reasoning} & \\

\cmidrule(lr){3-5}\cmidrule(lr){6-7}\cmidrule(lr){8-10}\cmidrule(lr){11-13} 
& & GU & OR & CD & TG & TP & GR & SL & PV & CA & CO & HR &  \\
\midrule

\rowcolor{lightdark} \multicolumn{14}{c}{\textbf{Based on Proprietary MLLMs}} \\
VideoAgent & -
& 76.67 & 41.21 & 43.80
& 32.92 & 53.89
& 38.19 & 52.66 & 64.85
& 86.67 & 91.84 & 86.08
& 51.53 \\

LLoVi & -
& 91.11 & 63.03 & 48.80
& 47.50 & 66.11
& 64.50 & 71.18 & 72.50
& 93.33 & 92.00 & 86.25
& 65.30 \\

VideoTree & -
& 84.44 & 49.70 & 51.20
& 49.38 & 71.11
& 64.00 & 66.47 & 64.58
& 91.11 & 90.00 & 86.25
& 62.30 \\

\rowcolor{lightbrown}
\multicolumn{14}{c}{\textbf{Open-source MLLMs Trained on Our ReVA Video Data}} \\

VideoLLaMA2 & 7B
& \best{98.89} & 76.97 & 70.00
& 59.06 & 78.33
& \secondbest{76.00} & \secondbest{82.94} & \secondbest{80.83}
& \secondbest{94.44} & \secondbest{92.00} & \best{90.00}
& 76.90 \\

LLaVA-NeXT-Video & 7B
& \secondbest{97.78} & 76.36 & \secondbest{70.40}
& 63.44 & \secondbest{81.67}
& 75.00 & 80.00 & 80.00
& \best{96.67} & \secondbest{92.00} & 86.25
& \secondbest{77.30} \\

VideoChat-Flash & 7B
& \best{98.89} & \best{82.42} & 64.80
& \secondbest{63.75} & 75.56
& \secondbest{76.00} & 78.82 & 77.50
& \best{96.67} & \secondbest{92.00} & 85.00
& 76.80 \\

VideoLLaVA & 7B
& 91.11 & 70.54 & 48.57
& 51.03 & 57.26
& 69.61 & 78.98 & 41.30
& 90.36 & 89.23 & 75.29
& 66.70 \\

BIMBA & 7B
& 95.56 & \secondbest{79.09} & 58.00
& 55.94 & 70.56
& 69.00 & 77.06 & 77.50
& 92.22 & \secondbest{92.00} & 81.25
& 72.35 \\



\textbf{ReMoSense (Ours)} & 7B

& \secondbest{97.78} & 76.67 & \best{71.49}
& \best{69.91} & \best{82.78}
& \best{77.00} & \best{85.88} & \best{81.67}
& \best{96.67} & \best{94.00} & \secondbest{87.50}
& \textbf{79.63} \\

\rowcolor{lightblue}
\multicolumn{14}{c}{\textbf{Open-source MLLMs Trained on Larger Scale Instruction Video Data}} \\
VideoLLaMA2 & 7B
& 86.67 & 62.73 & 55.60
& 41.25 & 59.44
& 65.50 & 68.82 & 63.33
& \secondbest{94.44} & 90.00 & 81.25
& 62.90 \\

LLaVA-NeXT-Video & 7B
& 88.89 & 65.76 & 53.20
& 35.00 & 65.56
& 52.50 & 66.47 & 30.83
& 90.00 & 86.00 & 78.75
& 56.95 \\

VideoChat-Flash & 7B
& \secondbest{95.56} & 77.27 & 57.20
& 51.56 & 72.22
& 64.00 & 61.76 & 32.92
& \best{95.56} & 90.00 & 78.75
& 64.25 \\

VideoLLaVA & 7B
& 94.44 & 67.88 & 56.80
& 51.56 & 61.11
& 66.00 & \secondbest{77.65} & 48.33
& 90.00 & 90.00 & 75.00
& 64.60 \\

BIMBA & 7B 
& \best{97.78} & \secondbest{79.09} & 55.20
& 52.50 & 75.00
& 66.00 & 77.06 & 75.42
& 93.33 & \secondbest{92.00} & \best{87.50}
& 71.70 \\

MovieChat-OneVision & 7B
& \best{97.78} & \best{81.82} & 60.80
& \secondbest{57.50} & 68.89
& \secondbest{72.00} & 74.12 & 71.67
& \secondbest{94.44} & \secondbest{92.00} & 82.50
& \secondbest{72.85} \\

VideoMind & 7B
& 91.11 & 73.94 & 51.60
& 36.25 & 70.00
& 53.00 & 60.00 & 52.92
& 88.89 & 88.00 & 80.00
& 61.00 \\

VITAL & 7B
& 76.67 & 70.61 & 60.40
& 34.69 & 64.44
& 61.50 & 62.94 & 68.75
& 86.67 & \secondbest{92.00} & \secondbest{86.25}
& 63.40 \\

Tarsier2 & 7B
& 94.40 & 62.10 & 56.80
& 51.20 & 58.30
& 67.00 & 77.10 & \secondbest{78.30}
& 86.70 & 90.00 & 75.00
& 66.80 \\

AVATAR & 7B
& 52.22 & 48.18 & 46.80
& 35.00 & 55.00
& 55.00 & 46.47 & 49.58
& 84.44 & 90.00 & 83.75
& 51.50 \\

InternVL3 & 8B
& 94.44 & 78.48 & \secondbest{62.00}
& 43.75 & \best{74.44}
& 71.00 & 64.71 & 45.83
& \best{95.56} & 88.00 & 85.00
& 66.65 \\


\textbf{ReMoSense (Ours)} & 7B
& \secondbest{95.56} & 74.55 & \best{67.20}
& \best{58.13} & \secondbest{73.89}
& \best{77.00} & \best{84.71} & \best{81.25}
& \best{95.56} & \best{94.00} & \best{87.50}
& \best{75.75} \\
\bottomrule
\end{tabular}
}
\begin{tablenotes}[flushleft]
\scriptsize
\item * GU: General Understanding, OR: Object and Land Cover Recognition, CD: Change Detection, TG: Temporal Grounding, TP: Trend and Pattern, GR: Geometric Relation, SL: Structural Layout, PV: Perspective and Viewpoint, CA: Causation Reasoning, CO: Consequence Reasoning, and HR: Hypothetical Reasoning.
\end{tablenotes}
\vspace{-0.6\baselineskip}
\label{tab:reva_results}
\end{table*}

\begin{figure*}[t]
\centering
\vspace{-0.2\baselineskip}
\begin{overpic}[width=\linewidth]{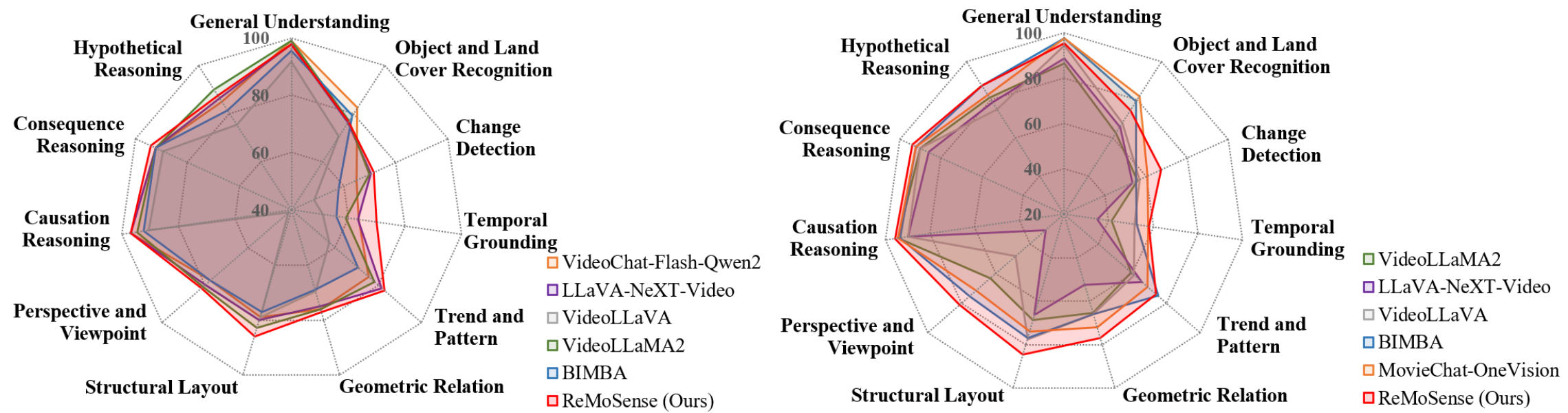}
  \put(2, -3){\small (a) Open-source MLLMs trained on ReVA}
  \put(48, -3){\small (b) Open-source MLLMs trained on large-scale data}
\end{overpic}
\vspace{0.2\baselineskip}
\caption{\textbf{Per-task comparisons with SOTA methods on ReVA validation set.}}
\vspace{-1\baselineskip}
\label{fig:radar_val}
\end{figure*}

\begin{table*}[t]
\centering
\caption{\textbf{Ablation studies of different inputs of ReMoSense on ReVA.}}
\label{tab:input_ablation}
\small
\begin{tabular}{lcccc}
\toprule
Category / Subcategory & Text-only & First frame & Middle frame & Full \\
\midrule
\textbf{Factual Perception}
& 29.55 & 38.96 & 39.33 & 75.08 \\
\quad General Understanding
& 35.00 & 22.78 & 24.44 & 97.78 \\
\quad Object and Land Cover Recognition
& 30.00 & 39.85 & 40.30 & 74.85 \\
\quad Change Detection
& 27.00 & 43.60 & 43.40 & 67.20 \\
\midrule
\textbf{Temporal Understanding}
& 27.00 & 37.70 & 37.80 & 64.40 \\
\quad Temporal Grounding
& 25.94 & 32.19 & 32.50 & 58.44 \\
\quad Trend and Pattern
& 28.89 & 47.50 & 47.22 & 75.00 \\
\midrule
\textbf{Spatial Reasoning}
& 28.52 & 37.13 & 37.13 & 81.31 \\
\quad Geometric Relation
& 29.00 & 38.00 & 37.50 & 78.00 \\
\quad Structural Layout
& 30.00 & 41.76 & 41.47 & 85.29 \\
\quad Perspective and Viewpoint
& 27.08 & 33.12 & 33.75 & 81.25 \\
\midrule
\textbf{Causal Reasoning}
& 41.82 & 72.73 & 73.18 & 94.55 \\
\quad Causation Reasoning
& 38.89 & 75.56 & 75.00 & 96.67 \\
\quad Consequence Reasoning
& 40.00 & 71.00 & 70.00 & 98.00 \\
\quad Hypothetical Reasoning
& 46.25 & 70.62 & 73.12 & 90.00 \\
\midrule
\textbf{Overall}
& 29.95 & 41.80 & 42.00 & 76.45 \\
\bottomrule
\end{tabular}
\end{table*}

\begin{figure*}[t]
\centering
\vspace{1\baselineskip}
\begin{overpic}[width=0.88\linewidth]{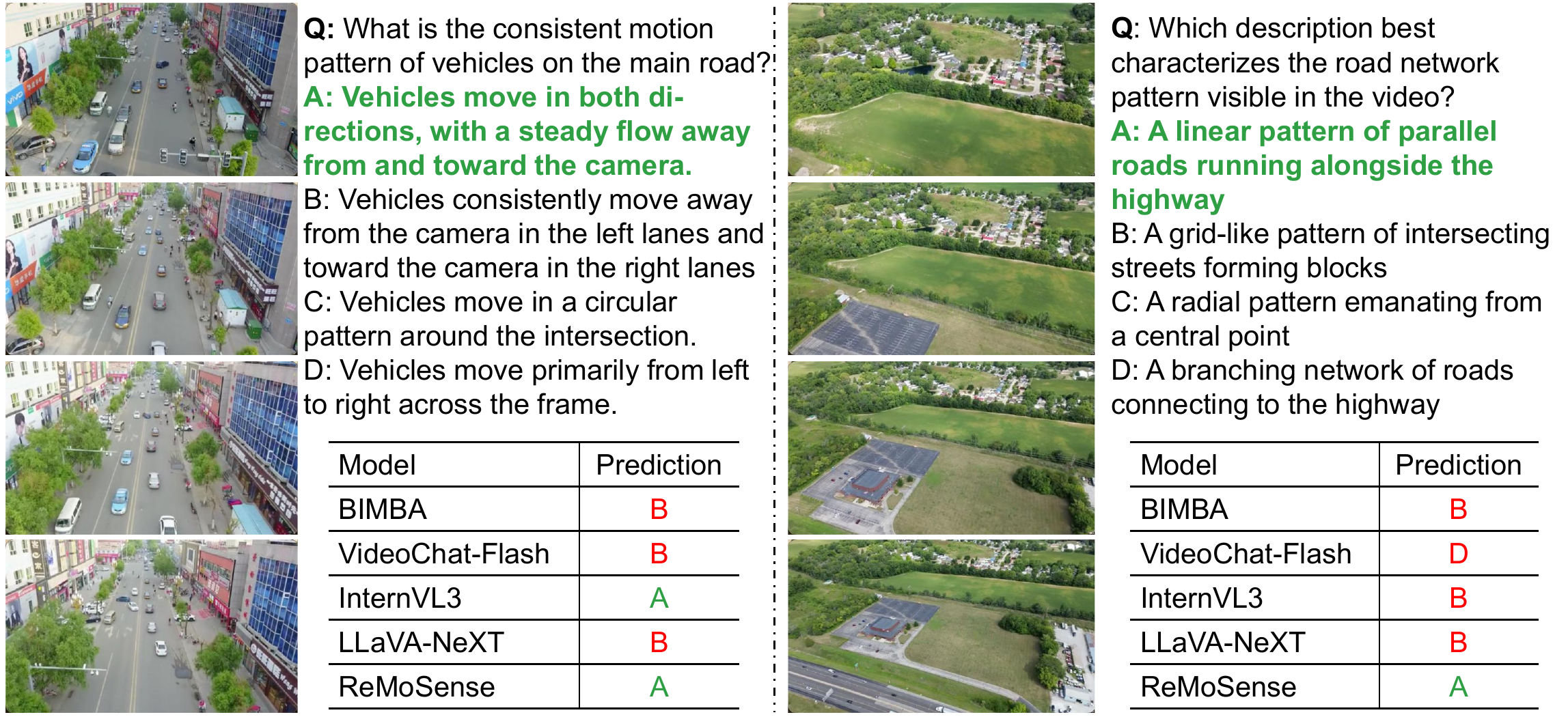}
  \put(9.5, -3.5){\small (a) Trend and Pattern QA}
  \put(62, -3.5){\small (b) Structural Layout QA}
\end{overpic}
\vspace{0.8\baselineskip}
\caption{\textbf{Visualization of answer predictions on ReVA.} The correct answers and correct predictions are highlighted in \textcolor{softgreen}{green}, and incorrect predictions are in \textcolor{red}{red}.}
\vspace{-1\baselineskip}
\label{fig:vis}
\end{figure*}

\section{Additional Results} \label{sec:addresult}
\subsection{Quantitative Results} \label{sec:quan}
Table \ref{tab:reva_results} summarizes the quantitative results on ReVA validation set. We evaluated 16 open-source and 3 proprietary MLLMs on 11 reasoning tasks. When fine-tuned on ReVA, our proposed method, ReMoSense, achieves 79.63\% overall accuracy, outperforming SOTA fine-tuned baseline LLaVA-NeXT-Video by 2.3\% and proprietary model LLoVi based on GPT-4o \cite{achiam2023gpt} by 14.3\%. Compared to our baseline model, BIMBA, the performance improves 7\%. When trained on large-scale data, ReMoSense achieves 75.75\% overall accuracy, outperforming MovieChat-OneVision by 3\% and proprietary model LLoVi by 10\%. Compared to our baseline model, BIMBA, the performance improves 4\%. \par

Among all categories, our model demonstrates the most improvements on Temporal Grounding (+6.5\%) and Structural Layout (+3\%), which we attribute to explicit motion modeling provided by ego motion alignment and object-level tracking. More quantitative results are in the Appendix. \par
  
\subsection{Ablation Studies} \label{seq:addabl}
To evaluate whether ReVA can be solved primarily through language priors or static visual cues, we evaluate ReMoSense on the ReVA test set under four input conditions: (1) QA pairs alone without video input (text-only); (2) QA pairs with only the first video frame; (3) QA pairs with only the middle video frame; and (4) the original full-video input. As shown in Table \ref{tab:input_ablation}, the text-only setting achieves only 29.95\% overall accuracy, while incorporating a single frame improves performance to 41.80\% and 42.00\% for the first- and middle-frame settings, respectively. In contrast, using the full video substantially increases the overall accuracy to 76.45\%, outperforming the two single-frame settings by 34\%. This improvement is consistent across all four categories. In particular, Temporal Understanding increases from 27.00\% with text only and approximately 37.8\% with a single frame to 64.40\% with the full video. More notably, despite the high full-video accuracy on Causal Reasoning (94.55\%), removing the video reduces performance to 41.82\%, while using a single frame achieves only 72\%. This progressive improvement indicates that generic linguistic or commonsense priors alone are insufficient for these questions: visual scene context provides substantial additional evidence, and the complete video provides information beyond an isolated frame. Together with our construction criterion that causal and hypothetical questions are conditioned on observable scene elements, these diagnostic results provide empirical support that these tasks are grounded in the remote-sensing video rather than primarily evaluating generic commonsense reasoning. \par

\subsection{Qualitative Results} \label{sec:qual}
We visualize qualitative results of our method with SOTA models using their default settings in Fig. \ref{fig:vis}. We present diverse scenarios from various tasks, including Trend and Pattern (column 1) and Structural Layout (column 2). Existing models struggle to handle complex relative motion pattern and locate remote layouts. For example, in column 1, these models cannot correctly identify the correct motion of cars. In column 2, the road structure pattern is misclassified as “grid”. In contrast, ReMoSense preserves object motion and layout integrity in crowded scenes, demonstrating superior robustness in challenging scenarios. \par

In addition, we provide the pseudo code for traing our proposed ReMoSense in Algorithm \ref{alg:remosense}. \par

\begin{algorithm}[t]
\caption{Training Process of ReMoSense}
\label{alg:remosense}
\begin{algorithmic}[1]
\Require Training set $\mathcal{D}=\{(V,Q,A)\}$, where 
$V=\{I^1,\dots,I^T\}$ is a video clip, $Q$ is the question, and 
$A$ is the ground-truth answer; image encoder $E_{\mathrm{img}}$; 
text encoder $E_{\mathrm{text}}$; Motion-Aware Alignment Module 
$\mathcal{M}_{\mathrm{align}}$; LLM decoder $\mathcal{F}_{\mathrm{LLM}}$.
\Ensure Trained ReMoSense model.

\For{each training sample $(V,Q,A)\in\mathcal{D}$}

    \State $X_q \leftarrow E_{\mathrm{text}}(Q)$
    \Comment{Encode question tokens}

    \For{$t=1$ to $T$}
        \State $X_{\mathrm{img}}^t \leftarrow E_{\mathrm{img}}(I^t)$
        \Comment{Patch-level visual tokens, 
        $X_{\mathrm{img}}^t\in\mathbb{R}^{H\times W\times C}$}
    \EndFor

    \For{$t=2$ to $T$}
        \State $\bar{X}_{\mathrm{img}}^{t-1}, \bar{X}_{\mathrm{img}}^{t}
        \leftarrow 
        \mathrm{Downsample}(X_{\mathrm{img}}^{t-1},X_{\mathrm{img}}^{t})$
        \Comment{Reduce resolution to $H/4\times W/4$}

        \State $\mathcal{C}^{t}(i,j) \leftarrow
        \frac{
        \bar{X}_{\mathrm{img}}^{t-1}(i)^\top
        \bar{X}_{\mathrm{img}}^{t}(j)}
        {
        \|\bar{X}_{\mathrm{img}}^{t-1}(i)\|_2
        \|\bar{X}_{\mathrm{img}}^{t}(j)\|_2}$
        \Comment{Cost volume between consecutive frames}

        \State $\mathcal{C}^{t\prime}
        \leftarrow \mathrm{MLP}(\mathcal{C}^{t})$
        \Comment{Correspondence representation}

        \State $X_{\mathrm{corr}}^{t}
        \leftarrow \mathrm{GlobalPool}(\mathcal{C}^{t\prime})$
        \Comment{Global correspondence token, 
        $X_{\mathrm{corr}}^{t}\in\mathbb{R}^{1\times C}$}
    \EndFor

    \State $X_{\mathrm{corr}}^{1}\leftarrow\mathbf{0}$
    \Comment{No previous frame for the first frame}

    \State Initialize $X_{\mathrm{obj}}^{1}
    \in\mathbb{R}^{K\times C}$
    \Comment{Learnable object-centric tokens}

    \State $Z^{1}\leftarrow
    \mathrm{Concat}
    (X_{\mathrm{img}}^{1},
    X_{\mathrm{corr}}^{1},
    X_{\mathrm{obj}}^{1})$
    \Comment{Initial frame representation}

    \For{$t=2$ to $T$}
        \State $X_{\mathrm{obj}}^{t}
        \leftarrow
        \mathrm{Transformer}
        \big(X_{\mathrm{obj}}^{t-1},X_{\mathrm{img}}^{t}\big)$
        \Comment{Update object tokens using current-frame features}

        \State $Z^{t}\leftarrow
        \mathrm{Concat}
        (X_{\mathrm{img}}^{t},
        X_{\mathrm{corr}}^{t},
        X_{\mathrm{obj}}^{t})$
        \Comment{Motion-enhanced frame representation}
    \EndFor

    \State $X^{1:T}
    \leftarrow
    \mathrm{Concat}(Z^{1},Z^{2},\dots,Z^{T})$
    \Comment{Motion-enhanced video representation}

    \State $X\leftarrow
    \mathrm{Concat}(X^{1:T},X_q)$
    \Comment{Multimodal input sequence}

    \State $\hat{A}\leftarrow
    \mathcal{F}_{\mathrm{LLM}}(X)$
    \Comment{Generate answer by next-token prediction}

    \State $\mathcal{L}\leftarrow
    \mathcal{L}_{\mathrm{CE}}(\hat{A},A)$
    \Comment{Language modeling loss}

    \State Update trainable parameters by gradient descent on $\mathcal{L}$

\EndFor

\State \Return Trained ReMoSense model
\end{algorithmic}
\end{algorithm}

\section{Additional Dataset Details} \label{sec:addataset}
\vspace{-1\baselineskip}
\noindent \textbf{Reviewing process.} Each QA pair is assigned one of four review labels during the annotation process:

\begin{itemize}
\item \textbf{Accept.} The question is clearly stated, the options are reasonable, and the labeled answer is correct.
\item \textbf{Modify.} The question is valid, but the provided answer is incorrect or missing from the options. In this case, annotators specify the correct answer and optionally provide revision notes.
\item \textbf{Reject.} The question is ambiguous, meaningless, duplicated, or cannot be answered using the video content.
\item \textbf{Review.} The QA pair is generally valid but requires additional review due to ambiguity, uncertainty, or domain-specific knowledge.
\end{itemize}

The eight first-stage reviewers were provided with the same annotation instructions and examples before verification. For each candidate QA pair, reviewers evaluated whether (1) the question is supported by the video, (2) the annotated answer is correct, (3) distractor options are plausible but incorrect, and (4) the wording is clear and unambiguous. For example, questions about object motion should focus on objects in the video rather than the camera itself, and questions that require object orientation are rejected if the orientation cannot be reliably inferred from the visual evidence. For problematic samples, annotators may provide correction suggestions or explanatory notes describing the issue and how the QA pair should be improved. These notes are later used to refine the final QA annotations. Samples requiring further review were then examined by two additional expert reviewers, who either finalized the corrected QA pair or removed it when a reliable answer could not be established from the video. \par

Table \ref{tab:category_annotations} shows the statistics of the annotation. This analysis shows how frequently the automatic pipeline produced directly acceptable samples and where stronger human intervention was required. Across all QA candidates, 73.0\% were accepted, 12.6\% required minor correction, 9.9\% rejected, and 4.5\% needed reviewing. In the meanwhile, \ref{tab:task_unaccepted} shows the task-wise number and ratio of the initial QA pairs that require revisions, which stands that temporal and spatial questions remain particularly challenging for the automatic QA generation pipeline. \par

\begin{table}[t]
\centering
\caption{Statistics of per-category annotations.}
\label{tab:category_annotations}
\setlength{\tabcolsep}{4pt}
\begin{tabular}{l|cccccc}
\toprule
\textbf{Category} & \makecell{\textbf{Initial}\\\textbf{candidates}} & \makecell{\textbf{Final}\\\textbf{total}} & \textbf{Accept}
& \makecell{\textbf{Minor}\\\textbf{correction}}
& \textbf{Reject} & \textbf{Review} \\
\midrule
Factual
& 6314
& 5773
& 4834 (76.6\%)
& 610 (9.7\%)
& 541 (8.6\%)
& 329 (5.2\%) \\

Temporal
& 5887
& 5072
& 3613 (61.4\%)
& 1058 (18.0\%)
& 815 (13.8\%)
& 401 (6.8\%) \\

Spatial
& 9849
& 9010
& 7389 (75.0\%)
& 1301 (13.2\%)
& 839 (8.5\%)
& 320 (3.2\%) \\

Causal
& 2113
& 1918
& 1797 (85.0\%)
& 72 (3.4\%)
& 195 (9.2\%)
& 49 (2.3\%) \\

\midrule
\textbf{Overall}
& \textbf{24163}
& \textbf{21773}
& \textbf{17633 (73.0\%)}
& \textbf{3041 (12.6\%)}
& \textbf{2390 (9.9\%)}
& \textbf{1099 (4.5\%)} \\
\bottomrule
\end{tabular}
\end{table}

\begin{table}[htbp]
\centering
\caption{Task-wise number and ratio of initial QA pairs that require revision.}
\vspace{-0.6\baselineskip}
\label{tab:task_unaccepted}
\begin{tabular}{lccc}
\toprule
\textbf{Categories} 
& \textbf{Count} 
& \makecell{\textbf{Required}\\\textbf{revision}}
& \textbf{Ratio} \\
\midrule
Change detection 
& 1879 & 774 & 41.2\% \\

Temporal grounding 
& 3142 & 1247 & 39.7\% \\

Trend and pattern analysis 
& 2745 & 1027 & 37.4\% \\

Viewpoint and perspective analysis 
& 2820 & 857 & 30.4\% \\

Geometric relation reasoning 
& 4194 & 1254 & 29.9\% \\

Object and land-cover recognition 
& 3014 & 588 & 19.5\% \\

Hypothetical reasoning 
& 701 & 123 & 17.5\% \\

Consequence reasoning 
& 704 & 111 & 15.8\% \\

Structural layout reasoning 
& 2835 & 349 & 12.3\% \\

Causation reasoning 
& 708 & 82 & 11.6\% \\

General Understanding 
& 1421 & 118 & 8.3\% \\

\bottomrule
\end{tabular}
\end{table}

Through this multi-step reviewing and correction process, we significantly reduce LLM hallucinations and ensure that the final QA pairs are accurate, meaningful, and suitable for evaluating video understanding in remote sensing. \par

\begin{table}[t]
\centering
\caption{Correct answer distribution before and after answer-option shuffling.}
\vspace{-0.6\baselineskip}
\label{tab:answer_bias}
\begin{tabular}{>{\centering\arraybackslash}p{1.4cm}
                >{\centering\arraybackslash}p{3.2cm}
                >{\centering\arraybackslash}p{2.8cm}}
\toprule
\textbf{Option} & \textbf{Initial generation} & \textbf{Final release} \\
\midrule
A & 15345 (70.5\%) & 5,443 (25.0\%) \\
B & 3730 (17.1\%) & 5,443 (25.0\%) \\
C & 1806 (8.3\%)  & 5,443 (25.0\%) \\
D & 802 (3.7\%)   & 5,444 (25.0\%) \\
E & 90 (0.4\%)    & 0 (0.0\%) \\
\bottomrule
\end{tabular}
\vspace{-0.6\baselineskip}
\end{table}

\begin{table}[ht!]
\centering
\caption{Agreement Results Across Different Categories}
\vspace{-0.6\baselineskip}
\label{tab:agreement}
\resizebox{0.8\textwidth}{!}{
\begin{tabular}{lcc}
\toprule
\textbf{Category} 
& \textbf{Pairwise agreement} 
& \textbf{Majority-vote agreement} \\
\midrule
Factual perception            & 1512 (84.4\%) & 90.6\% \\
Temporal understanding        & 1378 (86.3\%) & 92.1\% \\
Spatial \& viewpoint reasoning & 852 (82.2\%)  & 89.2\% \\
Causal reasoning              & 929 (79.0\%)  & 87.2\% \\
\midrule
\textbf{Overall}              & \textbf{4671 (83.4\%)} & \textbf{90.1\%} \\
\bottomrule
\end{tabular}
}
\end{table}

\begin{figure*}[!t]
\centering
\includegraphics[width=0.88\linewidth]{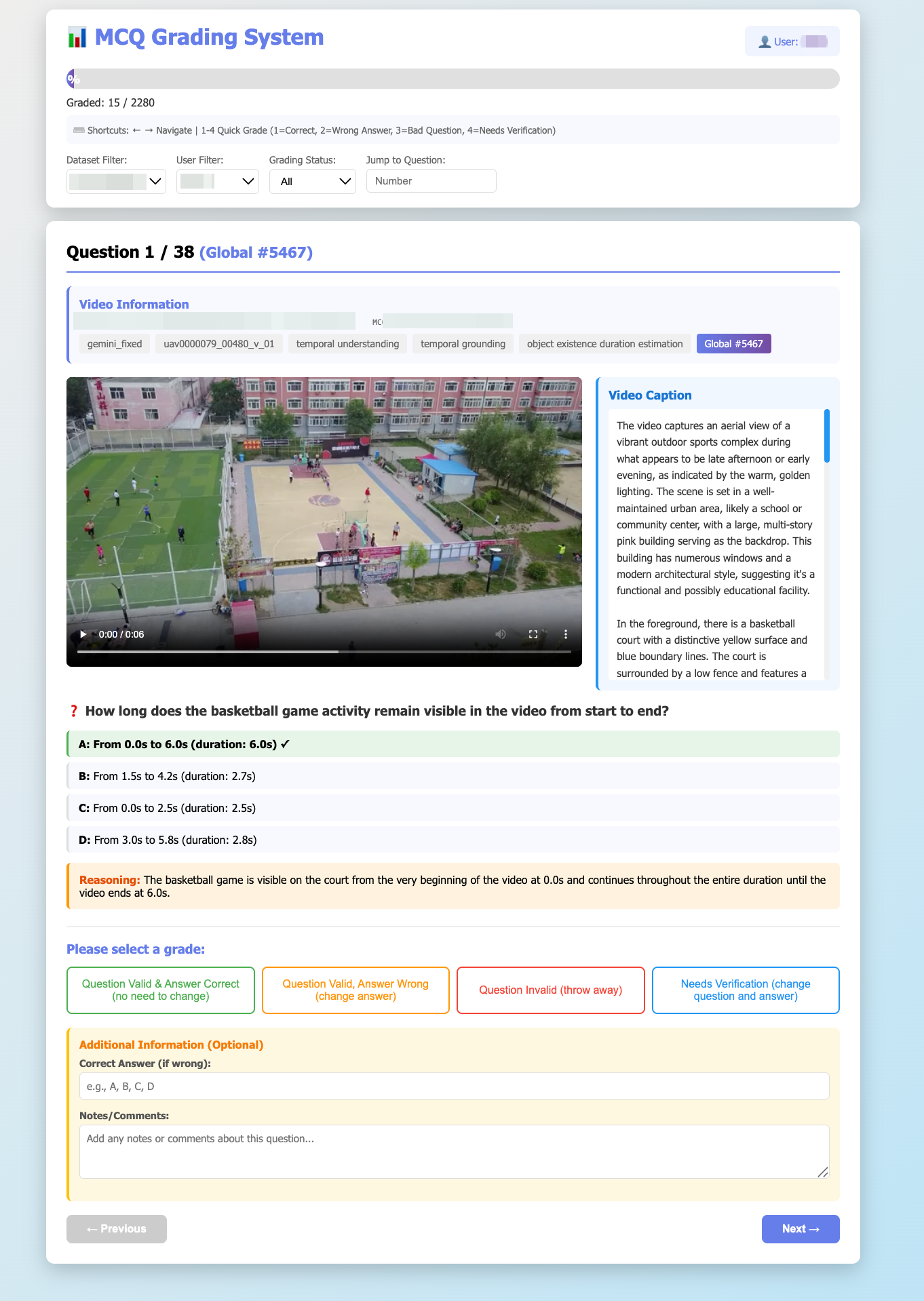}
\caption{\textbf{Web-based QA review platform} for human verification of generated QA pairs.}
\label{fig:platform}
\end{figure*}

\noindent \textbf{Answer Distribution and Bias Mitigation.} We observe a systematic answer-option bias in the initially generated question-answer pairs: the correct answer is assigned to option A in approximately 70.5\% of cases, while other options appear significantly less frequently. This bias originates from the generative models that tend to place the correct answer first. This imbalance leads to an accuracy of 70\% when always predicting A, further distorting model performance. \par

To mitigate this bias, we randomly shuffle the order of answer candidates for each question. This procedure ensures that the correct answers are evenly distributed across different option positions, preventing models from exploiting positional biases. After shuffling, as shown in Table \ref{tab:answer_bias}, the correct answer distribution becomes nearly uniform across options. \par

\begin{table*}[t]
\centering
\caption{\textbf{ReVA Question Type Details.} Each category has 2-3 QA tasks with representative question examples.}
\vspace{-0.6\baselineskip}
\small
\setlength{\tabcolsep}{6pt}
\resizebox{\textwidth}{!}{
\begin{tabular}{
>{\centering\arraybackslash}m{3cm}
>{\raggedright\arraybackslash}p{5.2cm}
>{\raggedright\arraybackslash}p{6.5cm}}
\toprule
\textbf{Category} & \textbf{QA task} & \textbf{Question example} \\
\midrule

\multirow{10}{*}{\textbf{Factual Perception}} & & Activity Recognition; \\
& & Object Existence Identification; \\
& General Understanding & Overall Scene Description; \\
\cmidrule(lr){2-3}
& & Object Counting; \\
& & Object Category Classification; \\
& Object and Land Cover Recognition & Object Spatial Distribution \\
\cmidrule(lr){2-3}
& & Object Appearance/Disappearance; \\
& & Shape Expansion/Shrinkage; \\
& Change Detection & Object Movement Detection \\
\midrule

\multirow{4}{*}{\centering\textbf{\makecell{Temporal\\Understanding}}} & & Object Appearance Time Estimation; \\
& Temporal Grounding & Object Existence Duration Estimation \\
\cmidrule(lr){2-3}
& & Motion Direction Estimation for Objects; \\
& Trend and Pattern & Periodic Change Detection \\
\midrule

\multirow{8}{*}{\textbf{Spatial Reasoning}} & & Relative Position Reasoning; \\
& & Object Orientation Estimation; \\
& Geometric Relation & Distance and Adjacency Estimation \\
\cmidrule(lr){2-3}
& & Spatial Arrangement Pattern Classification; \\
& Structural Layout & Road/River Network Pattern Recognition \\
\cmidrule(lr){2-3}
& & Camera Viewpoint Classification; \\
& Perspective and Viewpoint & Camera Motion State Recognition \\
\midrule

\multirow{4}{*}{\textbf{Causal Reasoning}}
& Causation Reasoning
& Cause Inference \\
\cmidrule(lr){2-3}
& Consequence Reasoning
& Future Outcome Prediction \\
\cmidrule(lr){2-3}
& Hypothetical Reasoning
& Counterfactual Inference \\
\bottomrule
\end{tabular}
}
\label{tab:taxonomy}
\vspace{-1.0\baselineskip}
\end{table*}

\noindent \textbf{Temporal grounding annotation.}
Temporal grounding annotations are created directly from the final video clips, rather than from the 32 uniformly sampled frames used for model inference. Annotators can continuously play, pause, and inspect the complete video to identify when the queried object or event becomes visible and when it disappears. Since the clips are approximately 15 seconds long, temporal intervals at a 0.5-second granularity can be reliably inspected by human annotators. The reported 32-frame sampling is applied only to model inputs during evaluation and does not constrain the temporal resolution of human annotation. Temporal Grounding is evaluated as a multiple-choice task, where models select the correct annotated interval rather than predicting frame-level temporal boundaries. \par

\noindent \textbf{Inter-reviewer agreement.}
To evaluate the reliability of our dataset annotations, we conduct an inter-reviewer agreement analysis among 200 questions in total. We randomly select questions across four major categories: 64 factual perception questions, 57 temporal understanding questions, 37 spatial and viewpoint reasoning, and 42 for causal reasoning. Eight reviewers independently re-annotated the 200-question audit set. For each question, we compare all possible annotator pairs ($C^8_2=28$ pairs per question), resulting in 5600 total pairwise comparisons across 200 questions. For those QA pairs that are marked as “Need review”, we recruited two additional experts to perform final adjudication of these cases (QA counted when only both experts reached agreement), resulting in a total of ten annotators involved in the complete verification pipeline.  \par

The overall pairwise agreement rate is 83.4\% (4671 agreements out of 5600 comparisons) and the majority-vote agreement is 90.1\%, indicating good annotation consistency. In addition, agreement varies slightly across question categories. As shown in Table \ref{tab:agreement}, temporal understanding shows the highest agreement, which is 86.3\% in pairwise agreement and 92.1\% in majority-vote agreement while casual reasoning shows the lowest agreement, it have 79.0\% pairwise agreement and 87.2\% majority-vote agreement. \par

\noindent \textbf{Review platform.} As shown in Fig. \ref{fig:platform}, we develop a web-based review platform that allows reviewers to simultaneously inspect the video, captions, and multiple-choice QA pairs. The interface displays the video player alongside automatically generated captions and the associated question with answer options, enabling annotators to verify whether the question is grounded in the visual evidence. \par

\section{ReVA Question Type Details} \label{sec:question_type_detail}
As shown in Table \ref{tab:taxonomy}, we present the question examples for 11 QA tasks under 4 categories. During dataset construction, the agent will generate question for each given example, and then decide whether to delete or modify it. \par

\section{Future Work} \label{sec:future}
As future work, we plan to extend ReVA beyond the current 15-second clips to long videos that can be used to evaluate long-horizon dependencies, such as multi-stage event evolution. The present clip length was a deliberate design choice to balance annotation quality, question difficulty, and computational efficiency. However, certain remote sensing scenarios unfold over longer time spans (e.g., slow construction progress, traffic pattern transitions, which are only partially reflected in short clips. In the future work, we will curate longer sequences, and introduce question types that require reasoning over extended temporal context. This extension will complement the current benchmark and provide a more comprehensive assessment of long-range spatiotemporal understanding in remote sensing videos. \par

\section{Ethics Statement} \label{sec:privacy}
Our UAV data collection follows a privacy-first policy and relevant flight regulations. Our self-collected videos were recorded only in rural areas with no visible people, while urban videos were obtained from existing publicly released datasets under their original licenses. We selected only legally accessible and authorized locations, avoided prohibited and sensitive areas, registered our UAV with the Federal Aviation Administration (FAA), completed the required TRUST training, and operated under applicable FAA DroneZone authorization and relevant state and local regulations. \par

We apply privacy screening throughout the pre-, during-, and post-review stages. Before annotation, every video is manually examined and clips containing identifiable individuals, visible faces, license plates, private property at close range, or other sensitive or personally identifiable information are excluded. During annotation and verification, reviewers are instructed to report any potential privacy issue missed during the initial screening. After annotation, videos with QA pairs are re-examined through cross-review, and any clip that does not satisfy our privacy criteria is removed. The same privacy screening is applied to third-party ERA and VisDrone videos, even though these datasets were previously publicly released. \par

Before final release, we also consulted legal counsel regarding dataset release and usage. We remove location-sensitive metadata, document the filtering criteria, intended research use, and licensing conditions in the dataset card, and provide a channel for reporting potential privacy concerns. Reported content will be promptly reviewed and, when appropriate, removed from future versions of the dataset.

\section{LLM Usage} \label{sec:llm}
In accordance with the applicable policy on the use of LLMs, we disclose that LLMs were used in two aspects of this work. First, during dataset construction, LLMs were used to generate initial question-answer pairs, which were subsequently carefully reviewed, revised, and verified by human annotators before inclusion in the final dataset. Second, LLMs were used as general-purpose writing tools to improve the clarity and readability of the manuscript. All research design, experimental analysis, result interpretation, and final decisions were conducted by the authors. \par

\section{Prompt Details} \label{sec:prompt_detail}
In this section, we present the complete set of prompts utilized in the data generation pipeline, alongside those employed for subjective evaluation. Specifically, these include caption generation prompt in Fig. \ref{fig:capgenerate}, caption consolidation prompt in Fig. \ref{fig:capconsolidate}, question-answer generation prompt in Fig. \ref{fig:answergenerate}, and QA refinement prompt in Fig. \ref{fig:qarefine}. \par

\begin{center}
\begin{minipage}{\linewidth}

\begin{minipage}{\linewidth}
\centering
\begin{tcolorbox}[
colback=gray!10,
colframe=gray!50,
boxrule=0.35pt,
arc=1pt,
left=0pt,right=0pt,top=0pt,bottom=0pt,
before skip=30pt,
after skip=5pt]

\centering
{\bfseries\normalsize Step 1: Caption Generation}

\raggedright
{\footnotesize
\textbf{Input:} <Video clip>

\medskip
\textbf{Prompt:}

Provide a detailed and objective description of the video. Focus only on information directly observable in the video and avoid speculative interpretations.

\medskip
Describe the content in the following order:

(1) Scene and environment \\
(2) Main objects or entities \\
(3) Actions and interactions \\
(4) Temporal progression of events \\
(5) Notable visual details

\medskip
Ensure that the description follows the chronological order of events
and contains approximately 100--200 words.
}

\medskip

\textbf{Output:} <Detailed caption describing the visual content of the video>
\end{tcolorbox}
\captionof{figure}{\textbf{Caption Generation Prompt.}}
\label{fig:capgenerate}
\end{minipage}
\begin{minipage}{\linewidth}
\begin{tcolorbox}[
colback=gray!10,
colframe=gray!50,
boxrule=0.35pt,
arc=1pt,
left=4pt,right=4pt,top=4pt,bottom=3pt,
before skip=30pt,
after skip=5pt
]

\centering
{\bfseries\normalsize Step 2: Caption Consolidation}

\raggedright
{\footnotesize
\textbf{Input:} <Video clip>, <Caption 1>, <Caption 2>, $\dots$, <Caption N>

\medskip
\textbf{Prompt:}

Produce a single comprehensive and coherent description of the video
by consolidating the information from the provided captions while
verifying their consistency with the visual content of the video.

\medskip
When generating the consolidated caption:

(1) Carefully review all provided captions \\
(2) Combine complementary information from different descriptions \\
(3) Remove redundant or contradictory content \\
(4) Preserve all important visual details and events \\
(5) Organize the description in a clear chronological order

\medskip
The captions are: <Caption 1>, <Caption 2>, $\dots$, <Caption N>

\medskip

\textbf{Output:} <Consolidated Caption>
}
\end{tcolorbox}
\captionof{figure}{\textbf{Caption Consolidation Prompt.}}
\label{fig:capconsolidate}
\end{minipage}

\end{minipage}

\end{center}

\begin{center}
\begin{minipage}{\linewidth}
\begin{minipage}{\linewidth}
\begin{tcolorbox}[
colback=gray!10,
colframe=gray!50,
boxrule=0.35pt,
arc=1pt,
left=4pt,right=4pt,top=3pt,bottom=3pt,
before skip=30pt,
after skip=5pt]

\centering
{\bfseries\normalsize Step 3: Question-Answer Generation}

\raggedright
{\footnotesize
\textbf{Input:} <Video clip>, <Consolidated Caption>

\medskip
\textbf{Prompt:}

Based on \{keywords\}, review the video and generate
\{questions\_per\_example\} multiple-choice questions about the
following type: \{QA task\}.

\medskip
Example questions for this category: \{examples\}

\medskip
Video description: <Consolidated Caption>

\medskip
When generating the questions:

(1) Ensure each question is grounded in the visual content of the video \\
(2) Focus on the specified reasoning category \\
(3) Avoid ambiguous or subjective questions \\
(4) Ensure that the correct answer can be inferred from the video content \\
(5) Generate four answer options, including one correct answer and multiple false options \\
(6) Provide a short reasoning rationale about how the correct answer is obtained

\medskip
\textbf{Output format:} <JSON of question-answer pairs>
}

\end{tcolorbox}
\captionof{figure}{\textbf{Question-Answer Generation Prompt.}}
\label{fig:answergenerate}
\end{minipage}

\begin{minipage}{\linewidth}
\begin{tcolorbox}[
colback=gray!10,
colframe=gray!50,
boxrule=0.35pt,
arc=1pt,
left=4pt,right=4pt,top=3pt,bottom=3pt,
before skip=30pt,
after skip=5pt]

\centering
{\bfseries\normalsize Step 4: Question-Answer Verification and Refinement}

\raggedright
{\footnotesize
\textbf{Input:} <Video Clip>, <Consolidated Caption>, <JSON of question-answer pairs>

\medskip
\textbf{Prompt:}

Review the video and verify the correct answer to the following
multiple-choice question.

\medskip
Question: \{question\}

\medskip
Options: \{options\}

\medskip
other candidate options: \{correct\_answer\}

\medskip
When verifying the answer:

(1) Ensure that the selected answer is fully supported by the visual content of the video \\
(2) Check whether other candidate options are consistent with the video \\
(3) Provide a concise rationale of why this answer is correct \\
(4) Avoid speculative interpretations beyond the observable content

\medskip
\textbf{Output format:} <JSON of refined question-answer pairs>

}
\end{tcolorbox}
\captionof{figure}{\textbf{Question-Answer Refinement Prompt.}}
\label{fig:qarefine}
\end{minipage}
\end{minipage}

\end{center}


\section{Examples} \label{sec:example}
\begin{figure*}[h]
\centering
\begin{overpic}[width=0.96\linewidth]{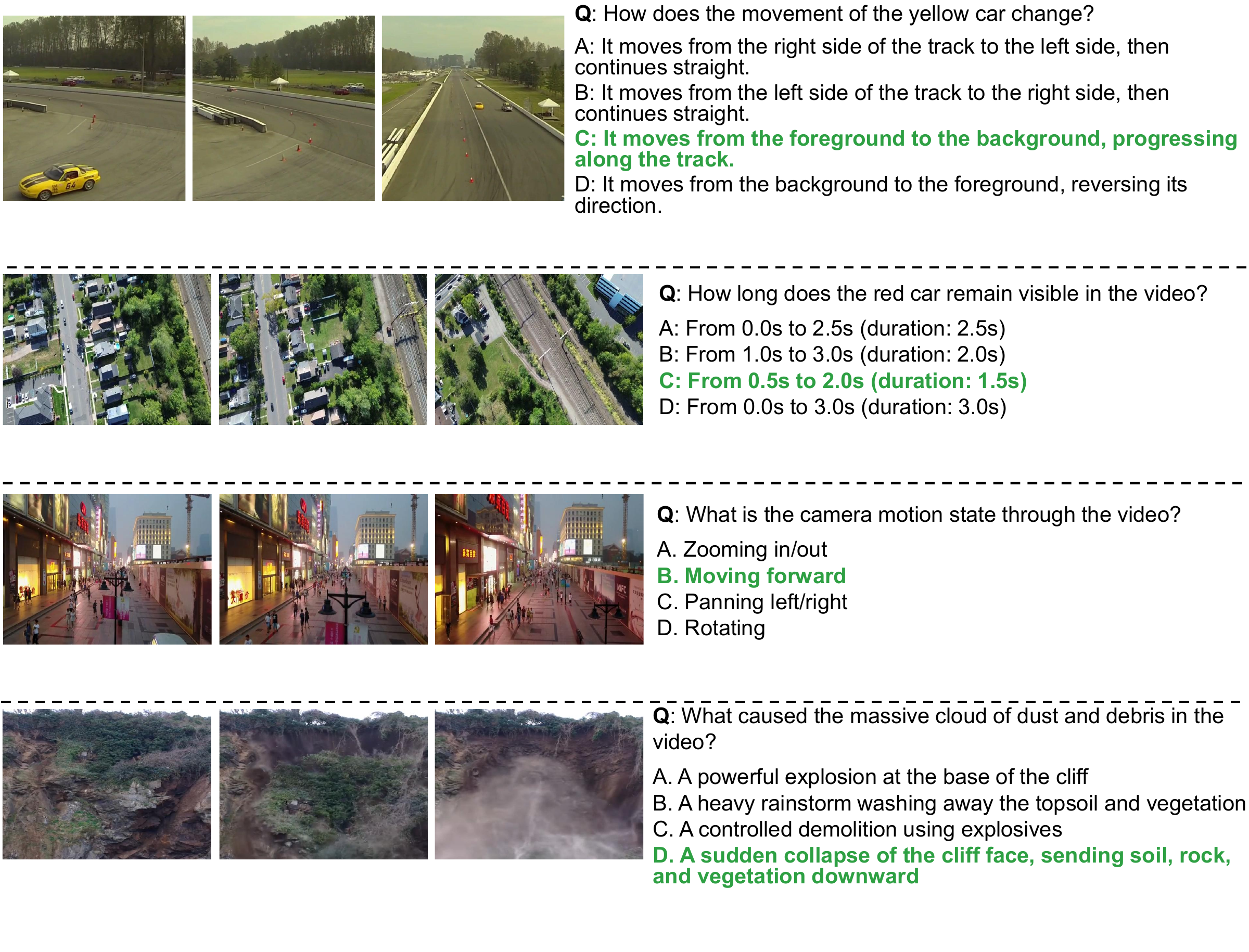}
  \put(38, 56.2){\small (a) Factual Perception}
  \put(35, 39){\small (b) Temporal Understanding}
  \put(38, 21.8){\small (c) Spatial Reasoning}
  \put(38, 3.2){\small (d) Causal Reasoning}
\end{overpic}
\vspace{-1.6\baselineskip}
\caption{\textbf{Examples of each category.} From top to bottom: (a) Factual Perception; (b) Temporal Understanding; (c) Spatial Reasoning; (d) Causal Reasoning.}
\label{fig:example_supp}
\vspace{0.6\baselineskip}
\end{figure*}
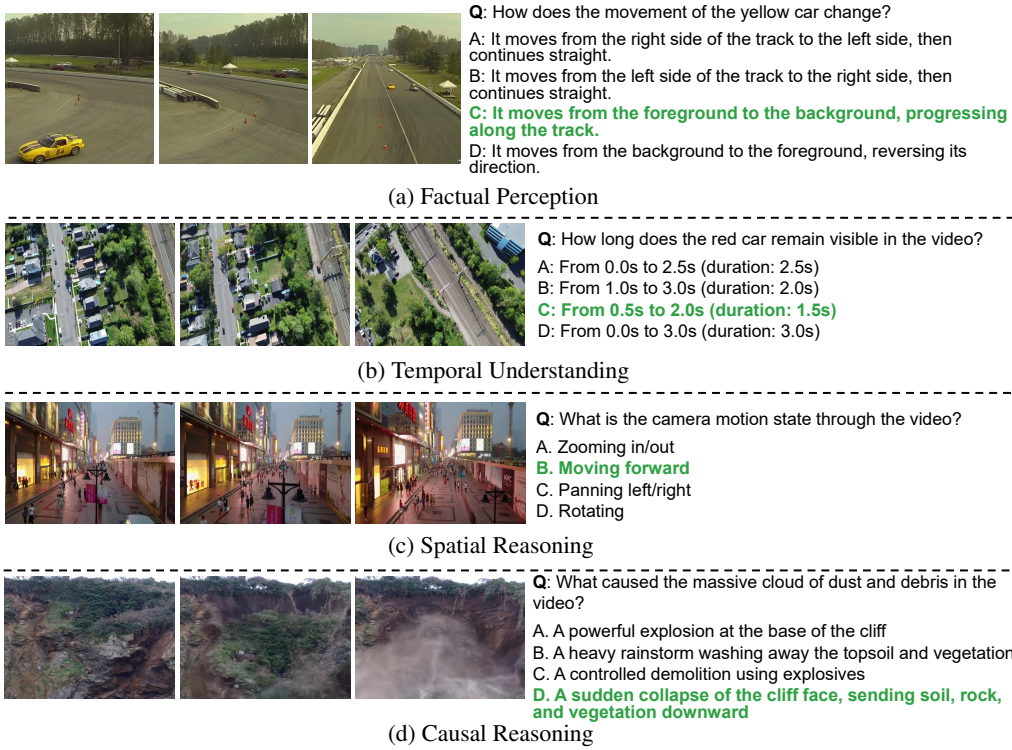

We further provide representative examples for each question category and QA task in our dataset. As illustrated in Fig. \ref{fig:example_supp}, each example consists of three video frames accompanied by a question and its corresponding answer, covering diverse scenes, resolutions, and temporal scales across the four major domains: Factual Perception, Temporal Understanding, Spatial Reasoning, and Causal Reasoning. \par

These examples illustrate both scene-level understanding and video-dependent reasoning in ReVA. Some factual and spatial questions may be answerable from a representative frame, while questions involving change, temporal grounding, trends, camera motion, or evolving events require evidence across multiple frames. For causal and hypothetical questions, we emphasize video-conditioned reasoning grounded in observable scene content, temporal evolution, spatial layouts, objects, and motion patterns, rather than unrestricted commonsense inference. We therefore view scene-level and video-dependent questions as complementary components of comprehensive remote sensing video understanding. \par

\clearpage

%% file: iclr2027_conference.bib
@article{liu2023visual,
  title={Visual instruction tuning},
  author={Liu, Haotian and Li, Chunyuan and Wu, Qingyang and Lee, Yong Jae},
  journal={Advances in neural information processing systems},
  volume={36},
  pages={34892--34916},
  year={2023}
}

@inproceedings{graumanEgo4DWorld30002022,
  title={Ego4d: Around the world in 3,000 hours of egocentric video},
  author={Grauman, Kristen and Westbury, Andrew and Byrne, Eugene and Chavis, Zachary and Furnari, Antonino and Girdhar, Rohit and Hamburger, Jackson and Jiang, Hao and Liu, Miao and Liu, Xingyu and others},
  booktitle={Proceedings of the IEEE/CVF conference on computer vision and pattern recognition},
  pages={18995--19012},
  year={2022}
}

@article{hu2025rsgpt,
  title={Rsgpt: A remote sensing vision language model and benchmark},
  author={Hu, Yuan and Yuan, Jianlong and Wen, Congcong and Lu, Xiaonan and Liu, Yu and Li, Xiang},
  journal={ISPRS Journal of Photogrammetry and Remote Sensing},
  volume={224},
  pages={272--286},
  year={2025},
  publisher={Elsevier}
}

@inproceedings{kuckreja2024geochat,
  title={Geochat: Grounded large vision-language model for remote sensing},
  author={Kuckreja, Kartik and Danish, Muhammad Sohail and Naseer, Muzammal and Das, Abhijit and Khan, Salman and Khan, Fahad Shahbaz},
  booktitle={Proceedings of the IEEE/CVF conference on computer vision and pattern recognition},
  pages={27831--27840},
  year={2024}
}

@article{li2024hrvqa,
  title={HRVQA: A Visual Question Answering benchmark for high-resolution aerial images},
  author={Li, Kun and Vosselman, George and Yang, Michael Ying},
  journal={ISPRS Journal of Photogrammetry and Remote Sensing},
  volume={214},
  pages={65--81},
  year={2024},
  publisher={Elsevier}
}

@article{li2024llava,
  title={Llava-onevision: Easy visual task transfer},
  author={Li, Bo and Zhang, Yuanhan and Guo, Dong and Zhang, Renrui and Li, Feng and Zhang, Hao and Zhang, Kaichen and Zhang, Peiyuan and Li, Yanwei and Liu, Ziwei and others},
  journal={arXiv preprint arXiv:2408.03326},
  year={2024}
}

@article{song2025moviechat,
  title={Moviechat+: Question-aware sparse memory for long video question answering},
  author={Song, Enxin and Chai, Wenhao and Ye, Tian and Hwang, Jenq-Neng and Li, Xi and Wang, Gaoang},
  journal={IEEE Transactions on Pattern Analysis and Machine Intelligence},
  year={2025},
  publisher={IEEE}
}

@article{wu2024longvideobench,
  title={Longvideobench: A benchmark for long-context interleaved video-language understanding},
  author={Wu, Haoning and Li, Dongxu and Chen, Bei and Li, Junnan},
  journal={Advances in Neural Information Processing Systems},
  volume={37},
  pages={28828--28857},
  year={2024}
}

@article{zhang2024llava,
  title={Llava-video: Video instruction tuning with synthetic data},
  author={Zhang, Yuanhan and Wu, Jinming and Li, Wei and Li, Bo and Ma, Zejun and Liu, Ziwei and Li, Chunyuan},
  journal={arXiv preprint arXiv:2410.02713},
  year={2024}
}

@inproceedings{zi2025rsvlm,
  title={RSVLM-QA: A Benchmark Dataset for Remote Sensing Vision Language Model-based Question Answering},
  author={Zi, Xing and Xiao, Jinghao and Shi, Yunxiao and Tao, Xian and Li, Jun and Braytee, Ali and Prasad, Mukesh},
  booktitle={Proceedings of the 33rd ACM International Conference on Multimedia},
  pages={12905--12911},
  year={2025}
}

@inproceedings{wang2024earthvqa,
  title={Earthvqa: Towards queryable earth via relational reasoning-based remote sensing visual question answering},
  author={Wang, Junjue and Zheng, Zhuo and Chen, Zihang and Ma, Ailong and Zhong, Yanfei},
  booktitle={Proceedings of the AAAI conference on artificial intelligence},
  volume={38},
  number={6},
  pages={5481--5489},
  year={2024}
}

@inproceedings{chappuis2022prompt,
  title={Prompt-RSVQA: Prompting visual context to a language model for remote sensing visual question answering},
  author={Chappuis, Christel and Zermatten, Val{\'e}rie and Lobry, Sylvain and Le Saux, Bertrand and Tuia, Devis},
  booktitle={Proceedings of the IEEE/CVF conference on computer vision and pattern recognition},
  pages={1372--1381},
  year={2022}
}

@article{zhang2023multistep,
  title={Multistep question-driven visual question answering for remote sensing},
  author={Zhang, Meimei and Chen, Fang and Li, Bin},
  journal={IEEE Transactions on Geoscience and Remote Sensing},
  volume={61},
  pages={1--12},
  year={2023},
  publisher={IEEE}
}

@article{yao2025remotereasoner,
  title={RemoteReasoner: Towards Unifying Geospatial Reasoning Workflow},
  author={Yao, Liang and Liu, Fan and Lu, Hongbo and Zhang, Chuanyi and Min, Rui and Xu, Shengxiang and Di, Shimin and Peng, Pai},
  journal={arXiv preprint arXiv:2507.19280},
  year={2025}
}

@article{massih2026reasoning,
  title={Reasoning with Pixel-level Precision: QVLM Architecture and SQuID Dataset for Quantitative Geospatial Analytics},
  author={Massih, Peter A and Cosatto, Eric},
  journal={arXiv preprint arXiv:2601.13401},
  year={2026}
}

@article{bashmal2023visual,
  title={Visual question generation from remote sensing images},
  author={Bashmal, Laila and Bazi, Yakoub and Melgani, Farid and Ricci, Riccardo and Al Rahhal, Mohamad M and Zuair, Mansour},
  journal={IEEE Journal of Selected Topics in Applied Earth Observations and Remote Sensing},
  volume={16},
  pages={3279--3293},
  year={2023},
  publisher={IEEE}
}

@article{zheng2021mutual,
  title={Mutual attention inception network for remote sensing visual question answering},
  author={Zheng, Xiangtao and Wang, Binqiang and Du, Xingqian and Lu, Xiaoqiang},
  journal={IEEE Transactions on Geoscience and Remote Sensing},
  volume={60},
  pages={1--14},
  year={2021},
  publisher={IEEE}
}

@article{rahnemoonfar2021floodnet,
  title={Floodnet: A high resolution aerial imagery dataset for post flood scene understanding},
  author={Rahnemoonfar, Maryam and Chowdhury, Tashnim and Sarkar, Argho and Varshney, Debvrat and Yari, Masoud and Murphy, Robin Roberson},
  journal={IEEE Access},
  volume={9},
  pages={89644--89654},
  year={2021},
  publisher={IEEE}
}

@inproceedings{yu2019activitynet,
  title={Activitynet-qa: A dataset for understanding complex web videos via question answering},
  author={Yu, Zhou and Xu, Dejing and Yu, Jun and Yu, Ting and Zhao, Zhou and Zhuang, Yueting and Tao, Dacheng},
  booktitle={Proceedings of the AAAI conference on artificial intelligence},
  volume={33},
  number={01},
  pages={9127--9134},
  year={2019}
}

@inproceedings{zadeh2019social,
  title={Social-iq: A question answering benchmark for artificial social intelligence},
  author={Zadeh, Amir and Chan, Michael and Liang, Paul Pu and Tong, Edmund and Morency, Louis-Philippe},
  booktitle={Proceedings of the IEEE/CVF Conference on Computer Vision and Pattern Recognition},
  pages={8807--8817},
  year={2019}
}

@inproceedings{xiao2021next,
  title={Next-qa: Next phase of question-answering to explaining temporal actions},
  author={Xiao, Junbin and Shang, Xindi and Yao, Angela and Chua, Tat-Seng},
  booktitle={Proceedings of the IEEE/CVF conference on computer vision and pattern recognition},
  pages={9777--9786},
  year={2021}
}

@inproceedings{castro2022wild,
  title={In-the-wild video question answering},
  author={Castro, Santiago and Deng, Naihao and Huang, Pingxuan and Burzo, Mihai and Mihalcea, Rada},
  booktitle={Proceedings of the 29th International Conference on Computational Linguistics},
  pages={5613--5635},
  year={2022}
}

@article{mangalam2023egoschema,
  title={Egoschema: A diagnostic benchmark for very long-form video language understanding},
  author={Mangalam, Karttikeya and Akshulakov, Raiymbek and Malik, Jitendra},
  journal={Advances in Neural Information Processing Systems},
  volume={36},
  pages={46212--46244},
  year={2023}
}

@inproceedings{li2024mvbench,
  title={Mvbench: A comprehensive multi-modal video understanding benchmark},
  author={Li, Kunchang and Wang, Yali and He, Yinan and Li, Yizhuo and Wang, Yi and Liu, Yi and Wang, Zun and Xu, Jilan and Chen, Guo and Luo, Ping and others},
  booktitle={Proceedings of the IEEE/CVF Conference on Computer Vision and Pattern Recognition},
  pages={22195--22206},
  year={2024}
}

@article{wu2024star,
  title={Star: A benchmark for situated reasoning in real-world videos},
  author={Wu, Bo and Yu, Shoubin and Chen, Zhenfang and Tenenbaum, Joshua B and Gan, Chuang},
  journal={arXiv preprint arXiv:2405.09711},
  year={2024}
}

@inproceedings{zhang2024simple,
  title={A simple llm framework for long-range video question-answering},
  author={Zhang, Ce and Lu, Taixi and Islam, Md Mohaiminul and Wang, Ziyang and Yu, Shoubin and Bansal, Mohit and Bertasius, Gedas},
  booktitle={Proceedings of the 2024 Conference on Empirical Methods in Natural Language Processing},
  pages={21715--21737},
  year={2024}
}

@inproceedings{wang2024videoagent,
  title={Videoagent: Long-form video understanding with large language model as agent},
  author={Wang, Xiaohan and Zhang, Yuhui and Zohar, Orr and Yeung-Levy, Serena},
  booktitle={European Conference on Computer Vision},
  pages={58--76},
  year={2024},
  organization={Springer}
}

@inproceedings{wang2025videotree,
  title={Videotree: Adaptive tree-based video representation for llm reasoning on long videos},
  author={Wang, Ziyang and Yu, Shoubin and Stengel-Eskin, Elias and Yoon, Jaehong and Cheng, Feng and Bertasius, Gedas and Bansal, Mohit},
  booktitle={Proceedings of the Computer Vision and Pattern Recognition Conference},
  pages={3272--3283},
  year={2025}
}

@inproceedings{islam2025bimba,
  title={Bimba: Selective-scan compression for long-range video question answering},
  author={Islam, Md Mohaiminul and Nagarajan, Tushar and Wang, Huiyu and Bertasius, Gedas and Torresani, Lorenzo},
  booktitle={Proceedings of the Computer Vision and Pattern Recognition Conference},
  pages={29096--29107},
  year={2025}
}

@article{li2024videochat,
  title={Videochat-flash: Hierarchical compression for long-context video modeling},
  author={Li, Xinhao and Wang, Yi and Yu, Jiashuo and Zeng, Xiangyu and Zhu, Yuhan and Huang, Haian and Gao, Jianfei and Li, Kunchang and He, Yinan and Wang, Chenting and others},
  journal={arXiv preprint arXiv:2501.00574},
  year={2024}
}

@misc{zhang2024llavanextvideo,
  title={LLaVA-NeXT: A Strong Zero-shot Video Understanding Model},
  url={https://llava-vl.github.io/blog/2024-04-30-llava-next-video/},
  author={Zhang, Yuanhan and Li, Bo and Liu, haotian and Lee, Yong jae and Gui, Liangke and Fu, Di and Feng, Jiashi and Liu, Ziwei and Li, Chunyuan},
  month={April},
  year={2024}
}

@inproceedings{lin2024video,
  title={Video-llava: Learning united visual representation by alignment before projection},
  author={Lin, Bin and Ye, Yang and Zhu, Bin and Cui, Jiaxi and Ning, Munan and Jin, Peng and Yuan, Li},
  booktitle={Proceedings of the 2024 conference on empirical methods in natural language processing},
  pages={5971--5984},
  year={2024}
}

@article{cheng2024videollama,
  title={Videollama 2: Advancing spatial-temporal modeling and audio understanding in video-llms},
  author={Cheng, Zesen and Leng, Sicong and Zhang, Hang and Xin, Yifei and Li, Xin and Chen, Guanzheng and Zhu, Yongxin and Zhang, Wenqi and Luo, Ziyang and Zhao, Deli and others},
  journal={arXiv preprint arXiv:2406.07476},
  year={2024}
}

@inproceedings{song2024moviechat,
  title={Moviechat: From dense token to sparse memory for long video understanding},
  author={Song, Enxin and Chai, Wenhao and Wang, Guanhong and Zhang, Yucheng and Zhou, Haoyang and Wu, Feiyang and Chi, Haozhe and Guo, Xun and Ye, Tian and Zhang, Yanting and others},
  booktitle={Proceedings of the IEEE/CVF Conference on Computer Vision and Pattern Recognition},
  pages={18221--18232},
  year={2024}
}

@article{bai2025qwen3,
  title={Qwen3-vl technical report},
  author={Bai, Shuai and Cai, Yuxuan and Chen, Ruizhe and Chen, Keqin and Chen, Xionghui and Cheng, Zesen and Deng, Lianghao and Ding, Wei and Gao, Chang and Ge, Chunjiang and others},
  journal={arXiv preprint arXiv:2511.21631},
  year={2025}
}

@article{zhu2025internvl3,
  title={Internvl3: Exploring advanced training and test-time recipes for open-source multimodal models},
  author={Zhu, Jinguo and Wang, Weiyun and Chen, Zhe and Liu, Zhaoyang and Ye, Shenglong and Gu, Lixin and Tian, Hao and Duan, Yuchen and Su, Weijie and Shao, Jie and others},
  journal={arXiv preprint arXiv:2504.10479},
  year={2025}
}

@article{loshchilov2017decoupled,
  title={Decoupled weight decay regularization},
  author={Loshchilov, Ilya and Hutter, Frank},
  journal={arXiv preprint arXiv:1711.05101},
  year={2017}
}

@inproceedings{rasley2020deepspeed,
  title={Deepspeed: System optimizations enable training deep learning models with over 100 billion parameters},
  author={Rasley, Jeff and Rajbhandari, Samyam and Ruwase, Olatunji and He, Yuxiong},
  booktitle={Proceedings of the 26th ACM SIGKDD international conference on knowledge discovery \& data mining},
  pages={3505--3506},
  year={2020}
}

@inproceedings{parikh2025roadsocial,
  title={Roadsocial: A diverse videoqa dataset and benchmark for road event understanding from social video narratives},
  author={Parikh, Chirag and Rawat, Deepti and Ghosh, Tathagata and Sarvadevabhatla, Ravi Kiran and others},
  booktitle={Proceedings of the Computer Vision and Pattern Recognition Conference},
  pages={19002--19011},
  year={2025}
}

@article{sarkar2023sam,
  title={Sam-vqa: Supervised attention-based visual question answering model for post-disaster damage assessment on remote sensing imagery},
  author={Sarkar, Argho and Chowdhury, Tashnim and Murphy, Robin Roberson and Gangopadhyay, Aryya and Rahnemoonfar, Maryam},
  journal={IEEE Transactions on Geoscience and Remote Sensing},
  volume={61},
  pages={1--16},
  year={2023},
  publisher={IEEE}
}

@inproceedings{yao2024cracknex,
  title={Cracknex: a few-shot low-light crack segmentation model based on retinex theory for uav inspections},
  author={Yao, Zhen and Xu, Jiawei and Hou, Shuhang and Chuah, Mooi Choo},
  booktitle={2024 IEEE International Conference on Robotics and Automation (ICRA)},
  pages={11155--11162},
  year={2024},
  organization={IEEE}
}

@article{mou2020era,
  title={Era: A data set and deep learning benchmark for event recognition in aerial videos [software and data sets]},
  author={Mou, Lichao and Hua, Yuansheng and Jin, Pu and Zhu, Xiao Xiang},
  journal={IEEE Geoscience and Remote Sensing Magazine},
  volume={8},
  number={4},
  pages={125--133},
  year={2020},
  publisher={IEEE}
}

@article{zhu2021detection,
  title={Detection and tracking meet drones challenge},
  author={Zhu, Pengfei and Wen, Longyin and Du, Dawei and Bian, Xiao and Fan, Heng and Hu, Qinghua and Ling, Haibin},
  journal={IEEE Transactions on Pattern Analysis and Machine Intelligence},
  volume={44},
  number={11},
  pages={7380--7399},
  year={2021},
  publisher={IEEE}
}

@article{wei2022chain,
  title={Chain-of-thought prompting elicits reasoning in large language models},
  author={Wei, Jason and Wang, Xuezhi and Schuurmans, Dale and Bosma, Maarten and Xia, Fei and Chi, Ed and Le, Quoc V and Zhou, Denny and others},
  journal={Advances in neural information processing systems},
  volume={35},
  pages={24824--24837},
  year={2022}
}

@article{dosovitskiy2020image,
  title={An image is worth 16x16 words: Transformers for image recognition at scale},
  author={Dosovitskiy, Alexey and Beyer, Lucas and Kolesnikov, Alexander and Weissenborn, Dirk and Zhai, Xiaohua and Unterthiner, Thomas and Dehghani, Mostafa and Minderer, Matthias and Heigold, Georg and Gelly, Sylvain and others},
  journal={arXiv preprint arXiv:2010.11929},
  year={2020}
}

@article{achiam2023gpt,
  title={Gpt-4 technical report},
  author={Achiam, Josh and Adler, Steven and Agarwal, Sandhini and Ahmad, Lama and Akkaya, Ilge and Aleman, Florencia Leoni and Almeida, Diogo and Altenschmidt, Janko and Altman, Sam and Anadkat, Shyamal and others},
  journal={arXiv preprint arXiv:2303.08774},
  year={2023}
}

@article{zhang2025thinking,
  title={Thinking With Videos: Multimodal Tool-Augmented Reinforcement Learning for Long Video Reasoning},
  author={Zhang, Haoji and Gu, Xin and Li, Jiawen and Ma, Chixiang and Bai, Sule and Zhang, Chubin and Zhang, Bowen and Zhou, Zhichao and He, Dongliang and Tang, Yansong},
  journal={arXiv preprint arXiv:2508.04416},
  year={2025}
}

@article{yuan2025tarsier2,
  title={Tarsier2: Advancing large vision-language models from detailed video description to comprehensive video understanding},
  author={Yuan, Liping and Wang, Jiawei and Sun, Haomiao and Zhang, Yuchen and Lin, Yuan},
  journal={arXiv preprint arXiv:2501.07888},
  year={2025}
}

@article{liu2025videomind,
  title={Videomind: A chain-of-lora agent for long video reasoning},
  author={Liu, Ye and Qinghong Lin, Kevin and Chen, Chang Wen and Shou, Mike Zheng},
  journal={arXiv e-prints},
  pages={arXiv--2503},
  year={2025}
}

@article{bai2023qwen,
  title={Qwen technical report},
  author={Bai, Jinze and Bai, Shuai and Chu, Yunfei and Cui, Zeyu and Dang, Kai and Deng, Xiaodong and Fan, Yang and Ge, Wenbin and Han, Yu and Huang, Fei and others},
  journal={arXiv preprint arXiv:2309.16609},
  year={2023}
}

@article{hu2022lora,
  title={Lora: Low-rank adaptation of large language models.},
  author={Hu, Edward J and Shen, Yelong and Wallis, Phillip and Allen-Zhu, Zeyuan and Li, Yuanzhi and Wang, Shean and Wang, Liang and Chen, Weizhu and others},
  journal={Iclr},
  volume={1},
  number={2},
  pages={3},
  year={2022}
}

@inproceedings{du2018unmanned,
  title={The unmanned aerial vehicle benchmark: Object detection and tracking},
  author={Du, Dawei and Qi, Yuankai and Yu, Hongyang and Yang, Yifan and Duan, Kaiwen and Li, Guorong and Zhang, Weigang and Huang, Qingming and Tian, Qi},
  booktitle={Proceedings of the European conference on computer vision (ECCV)},
  pages={370--386},
  year={2018}
}

@inproceedings{liu2025nvila,
  title={Nvila: Efficient frontier visual language models},
  author={Liu, Zhijian and Zhu, Ligeng and Shi, Baifeng and Zhang, Zhuoyang and Lou, Yuming and Yang, Shang and Xi, Haocheng and Cao, Shiyi and Gu, Yuxian and Li, Dacheng and others},
  booktitle={Proceedings of the IEEE/CVF Conference on Computer Vision and Pattern Recognition},
  pages={4122--4134},
  year={2025}
}

@article{ferrag2026uavbench,
  title={UAVBench: An open benchmark dataset for autonomous and agentic AI UAV systems via LLM-generated flight scenarios},
  author={Ferrag, Mohamed Amine and Lakas, Abderrahmane and Debbah, Merouane},
  journal={IEEE Open Journal of Vehicular Technology},
  year={2026},
  publisher={IEEE}
}

@article{zhou2026rsvideo,
  title={RSVideo: Are Your Vision-Language Models Ready for Remote Sensing Videos?},
  author={Zhou, Hongjie and Wang, Shiqin and Chen, Haoyang and Guo, Haonan and Wang, Di and Liu, Juhua and Lin, Fu and Luo, Yong},
  journal={arXiv preprint arXiv:2608.02039},
  year={2026}
}

@article{sun2026memory,
  title={Memory-Augmented Multimodal Large Language Models for Small Object Understanding in Streaming Aerial Videos},
  author={Sun, Penglei and Huang, Yehua and Tao, Zhuoli and Li, Xiang and Guan, Runwei and Song, Yaoxian and Zhao, Kaiyong and Ding, Henghui and Han, Bo and Yang, Yang and others},
  journal={arXiv preprint arXiv:2607.19857},
  year={2026}
}
